%% file: main.tex
\documentclass[accepted]{uai2021}
\usepackage[american]{babel}
\usepackage{xr}

\usepackage{natbib} 
\usepackage{mathtools} 
\usepackage{booktabs} 
\usepackage{tikz} 

\input{math_commands.tex}

\usepackage{microtype}
\usepackage{graphicx}
\usepackage{amsmath}
\usepackage{amsfonts}
\usepackage{amssymb}
\usepackage{mathrsfs}

\usepackage{xspace}
\usepackage{xcolor}
\usepackage{caption}
\usepackage{subcaption}
\usepackage{rotating}
\usepackage{makecell}
\usepackage[export]{adjustbox}

\usepackage{algpseudocode} 
\usepackage{algorithm} 

\definecolor{ourspecialtextcolor}{rgb}{0.528, 0.471, 0.701} 
\hypersetup{
    colorlinks = true,
    linkbordercolor = {white},
	linkcolor = black, 
	citecolor=ourspecialtextcolor, 
	urlcolor=blue
}

\makeatletter
\DeclareRobustCommand\onedot{\futurelet\@let@token\@onedot}
\def\@onedot{\ifx\@let@token.\else.\null\fi\xspace}
\def\eg{\emph{e.g}\onedot} 
\def\ie{\emph{i.e}\onedot} 
 
\def\etc{\emph{etc}\onedot} 
\def\wrt{w.r.t\onedot}

\makeatother

\newcommand{\ourmodel}{NP-DRAW}

\makeatletter
\newcommand\email[2][]%
   {\newaffiltrue\let\AB@blk@and\AB@pand
      \if\relax#1\relax\def\AB@note{\AB@thenote}\else\def\AB@note{\relax}%
        \setcounter{Maxaffil}{0}\fi
      \begingroup
        \let\protect\@unexpandable@protect
        \def\thanks{\protect\thanks}\def\footnote{\protect\footnote}%
        \@temptokena=\expandafter{\AB@authors}%
        {\def\\{\protect\\\protect\Affilfont}\xdef\AB@temp{#2}}%
         \xdef\AB@authors{\the\@temptokena\AB@las\AB@au@str
         \protect\\[\affilsep]\protect\Affilfont\AB@temp}%
         \gdef\AB@las{}\gdef\AB@au@str{}%
        {\def\\{, \ignorespaces}\xdef\AB@temp{#2}}%
        \@temptokena=\expandafter{\AB@affillist}%
        \xdef\AB@affillist{\the\@temptokena \AB@affilsep
          \AB@affilnote{}\protect\Affilfont\AB@temp}%
      \endgroup
       \let\AB@affilsep\AB@affilsepx
}

\makeatother

\title{NP-DRAW: A Non-Parametric Structured Latent Variable Model \\ for Image Generation}

\author[1,2]{{Xiaohui Zeng}}
\author[1,2,4]{Raquel Urtasun}
\author[1,2,4]{Richard Zemel}
\author[1,2,3]{Sanja Fidler}
\author[1,2]{Renjie Liao}
\affil[1]{University of Toronto,  ${}^2$ Vector Institute,  ${}^{3}$ NVIDIA, ${}^{4}$ Canadian Institute for Advanced Research } 
\email{\tt  \{xiaohui, urtasun, zemel, fidler,  rjliao\}@cs.toronto.edu}

\begin{document}
\maketitle

\begin{abstract}
In this paper, we present a non-parametric structured latent variable model for image generation, called NP-DRAW, which sequentially draws on a latent canvas in a part-by-part fashion and then decodes the image from the canvas.
Our key contributions are as follows. 
1) We propose a non-parametric prior distribution over the appearance of image parts so that the latent variable ``what-to-draw'' per step becomes a categorical random variable.
This improves the expressiveness and greatly eases the learning compared to Gaussians used in the literature.
2) We model the sequential dependency structure of parts via a Transformer, which is more powerful and easier to train compared to RNNs used in the literature.
3) We propose an effective heuristic parsing algorithm to pre-train the prior.
Experiments on MNIST, Omniglot, CIFAR-10, and CelebA show that our method significantly outperforms previous structured image models like DRAW and AIR and is competitive to other generic generative models.
Moreover, we show that our model's inherent compositionality and interpretability bring significant benefits in the low-data learning regime and latent space editing. Code is available at  

\url{https://github.com/ZENGXH/NPDRAW}. 
\end{abstract}

\input{intro.tex}

\input{related.tex}
\input{method.tex}

\input{exp.tex}

\input{conclusion.tex}

\begin{acknowledgements} 
This work was supported by NSERC and DARPA’s XAI program. SF acknowledges the Canada CIFAR AI Chair award at the Vector Institute. RL was supported by RBC Fellowship. We would like to thank Yang Song, Yujia Li, David Duvenaud, and anonymous reviewers for valuable feedback.
\end{acknowledgements}

\bibliography{main}
\clearpage

\input{arxiv_appen.tex}

\providecommand{\upGamma}{\Gamma}
\providecommand{\uppi}{\pi}

\end{document}

%% file: math_commands.tex
\usepackage{amsmath,amsfonts,bm}

\def\1{\bm{1}}

\def\rvc{{\mathbf{c}}}

\def\rvx{{\mathbf{x}}}

\def\rvz{{\mathbf{z}}}

\DeclareMathAlphabet{\mathsfit}{\encodingdefault}{\sfdefault}{m}{sl}
\SetMathAlphabet{\mathsfit}{bold}{\encodingdefault}{\sfdefault}{bx}{n}
\newcommand{\tens}[1]{\bm{\mathsfit{#1}}}

\def\tE{{\tens{E}}}

\newcommand{\E}{\mathbb{E}}

\newcommand{\KL}{D_{\mathrm{KL}}}



%% file: intro.tex
\section{Introduction}\label{sect:intro}

Humans understand and create images in a structural way.
Given a scene, one can quickly recognize different objects and their functional relationships.
While drawing a picture, humans naturally draw one part at a time and iteratively improve the drawing.
However, many existing generative models like variational autoencoders (VAEs) \citep{kingma2013auto} and generative adversarial networks (GANs) \citep{goodfellow2014generative} generate all pixels of an image simultaneously, which is unnatural and hardly interpretable.
For example, the latent space of VAEs is a high-dimensional vector space which is hard to visualize and meaningfully manipulate.
Therefore, building more human-like and interpretable models for image understanding and generation has been an important quest \citep{lake2015human}.
Note that good generative models could often facilitate both image understanding and generation (\eg, a good encoder in VAEs could recognize specific patterns in the image, whereas a good decoder could generate images with high quality).

In this paper, we propose a structured probabilistic model for image generation, which follows the line of Deep Recurrent Attentive Writer (DRAW) \citep{gregor2015draw} and Attend-Infer-Repeat (AIR) networks \citep{eslami2016attend} that mimic how humans draw in a part-by-part fashion.
In particular, we sequentially decide whether-to-draw, generate parts (what-to-draw) and their locations (where-to-draw), and draw them on a latent canvas from which the final image is decoded.
Our main contributions are as follows.

First, instead of using Gaussian latent variables as in DRAW and AIR which are too limited to capture the large variation of the appearance of the individual image parts, we resort to a non-parametric categorical distribution, which is more expressive, \eg, capturing multi-modalities in the appearance distribution.
We build a dictionary of exemplar parts by performing k-medoids clustering on random local patches collected from images.
This categorical distribution is then defined over the discrete choice of the exemplar parts.
By doing so, we convert the hard what-to-draw problem to an easier which-to-choose problem.

Second, instead of using recurrent neural networks (RNNs) to construct the encoder, prior, and decoder like what DRAW and AIR do, we simplify the overall model by using a Transformer-based prior and plain convolutional neural networks (CNNs) based encoder and decoder. 
Transformer excels at capturing dependencies among parts and is easier to train compared to RNNs since it does not carry hidden states over generation steps.
This design choice significantly eases the learning and makes our model more scalable.

At last, we propose a heuristic algorithm to parse the latent variables given an image so that we can effectively pre-train our prior network.
Moreover, at each generation step, we allow the model to decide whether-to-draw by sampling from a learned Bernoulli distribution.
If yes, then we draw the generated part in the generated location of the canvas. 
Otherwise, we skip this step and leave the canvas unchanged.
Therefore, although fixing the total number of generation steps, the effective number of steps varies from image to image, thus being more flexible.

We extensively benchmark our model on four datasets, \ie, MNIST, Omniglot, CIFAR-10, and CelebA.
In terms of the sample quality, our model is significantly superior to other structured generative models like DRAW \citep{gregor2015draw} and AIR \citep{eslami2016attend} and is comparable to state-of-the-art generic generative models like WGAN \citep{arjovsky2017wasserstein}.
Furthermore, under small portions of training data, our model consistently outperforms other structured generative models and is comparable to the best generic generative model. 
The strong performances under the low-data regime support that our model better exploits the compositionality in the data, since otherwise it would require much more training samples to reach satisfying performances.
We also demonstrate the interpretability of our model by manipulating sampled images in a compositional way through latent space editing.

The rest of the paper is organized as follows. 
In Section \ref{sect:relate_work}, we discuss the related work.
Then we introduce the details of our model in Section \ref{sect:model}.
Section \ref{sect:exp} provides experimental results.
Section \ref{sect:conclusion} concludes the paper.


%% file: related.tex
\section{Related Work}\label{sect:relate_work}

Generative models for images have been studied for several decades \citep{dayan1995helmholtz,zhu1997minimax,freeman2000learning,crandall2005spatial,jin2006context}.
In the context of deep learning, generative models for images have been extensively studied from many aspects or principles, \eg, 
deep belief nets \citep{hinton2006fast}, 
deep Boltzmann machines \citep{salakhutdinov2009deep},
variational auto-encoders (VAEs) \citep{kingma2013auto,rezende2014stochastic,vahdat2020nvae,razavi2019generating}, 
generative adversarial networks (GANs) \citep{goodfellow2014generative,arjovsky2017wasserstein,gulrajani2017improved,brock2018large}, 
generative moment matching networks \citep{li2015generative,dziugaite2015training},
normalizing flows \citep{rezende2015variational,dinh2016density,kingma2018glow,papamakarios2019normalizing}, 
energy based models (EBMs) \citep{du2019implicit,song2019generative}, 
and deep diffusion models \citep{sohl2015deep,ho2020denoising}.
Our work also relates to examplar VAE~\citep{norouzi2020exemplar} which use a non-parametric prior with exemplars sampled from the training set. 
However, most of the above generative models generate all pixel values of an image at once, which contrasts with how humans draw pictures.

In this paper, we focus on a class of deep generative models that sequentially generate an image in a part-by-part fashion.
One of the first such models is the Deep Recurrent Attention Writer (DRAW) \citep{gregor2015draw}.
It is a sequential variational auto-encoding framework with the encoder and the decoder parameterized by two recurrent neural networks (RNNs) respectively.
During inference, the encoder uses the spatial attention mechanism to decide where-to-read and then constructs the approximated posterior.
During generation, the decoder also leverages spatial attention to decide where-to-write and creates what-to-write.
Convolutional DRAW \citep{gregor2016towards} further improves DRAW by leveraging convolutions to construct the encoder/decoder and learning the prior.
Attend-Infer-Repeat (AIR) networks \citep{eslami2016attend} enriches this model class by additionally learning to choose the appropriate number of inference/generation steps.
AIR was later extended to a sequential version (SeqAIR) \citep{kosiorek2018sequential}, which models videos and can handle dynamically changing objects.
More recently, VQ-DRAW \citep{nichol2020vq} introduces vector quantization in the latent space and decides what-to-draw by choosing one of the proposals output by a network instead of learning a Gaussian.
Arguably, PixelCNN \citep{van2016conditional} can be viewed as an extreme case of this model class that generates one pixel at a time conditioned on previously generated ones without considering a latent space.
There are also stroke based generative models like SPIRAL~\citep{ganin2018synthesizing}, Cose~\citep{aksan2020cose}, and SketchEmbedNet~\citep{wang2020sketchembednet}. SPIRAL generates images through a sequence of strokes while Cose and SketchEmbedNet focus on generating sketch images.

%% file: method.tex
\section{Model}\label{sect:model}

\begin{figure*}[h]
    \begin{center} 
    \includegraphics[width=\textwidth]{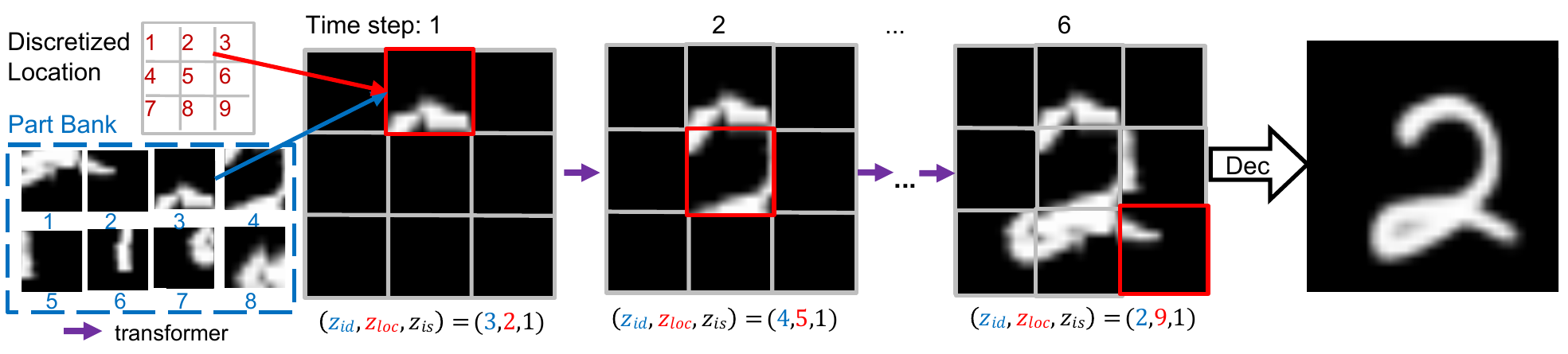} 
    \end{center}    
    \vspace{-0.5cm}
    \caption{The generative process of our model. At each time step, a Transformer-based auto-regressive prior generates latent variables including what-to-draw $\rvz_{\text{id}}$ (choosing a part from the part bank), where-to-draw $\rvz_{\text{loc}}$ (choosing one of the discretized locations), and whether-to-draw $\rvz_{\text{is}}$ (whether skipping the current step). The sampled image is generated from the final canvas via a decoder.}
    \vspace{-0.4cm}
    \label{fig:model}
\end{figure*}

In this section, we introduce our model \ourmodel, which follows the general framework of VAEs \citep{kingma2013auto}.
Our generative model $p_{\theta}(\rvx \vert \rvz) p_{\theta}(\rvz)$, parameterized by $\theta$, first generates some latent variables $\rvz$ from the prior $p_{\theta}(\rvz)$ and then generates an image $\rvx$ conditioned on $\rvz$ via the decoder $p_{\theta}(\rvx \vert \rvz)$.
Given an image $\rvx$, the inference of latent variables $\rvz$ is implemented by an amortized encoder network $q_{\phi}(\rvz \vert \rvx)$ parameterized by $\phi$.
Ideally, $\rvz$ should describe attributes of the image, \eg, appearances and structural relationships of objects and parts.  
Many existing latent variable models, \eg, VAEs, use a single vector $\rvz$, which hardly captures such structural information.

\paragraph{Part-based Image Representations} 

Inspired by part-based models in image and object modeling \citep{hinton1979some,fischler1973representation,felzenszwalb2005pictorial,zhu2007stochastic}, we introduce multiple groups of latent variables $\rvz = \{\rvz^{1}, \dots, \rvz^{T}\}$ where each group $\rvz^{t}$ corresponds to the drawing of a part of the image and $T$ is the total number of generation steps.
Here we use parts to refer to local spatial regions of images. 
They could be semantically meaningful parts of objects like wheels of cars or some recurring regions of the scene.
In particular, at the $t$-th generation step, the group $\rvz^{t}$ describes an image part in terms of its location $\rvz_{\text{loc}}^{t}$, its appearance $\rvz_{\text{id}}^{t}$, and whether we draw it $\rvz_{\text{is}}^{t}$ on the latent canvas $\rvc^{t}$, \ie, $\rvz^{t} = [\rvz_{\text{loc}}^{t}, \rvz_{\text{id}}^{t}, \rvz_{\text{is}}^{t}]$.
Although we fixed $T$ as a hyperparameter following DRAW, we allow the model to learn to skip a step (\ie, $\rvz_{\text{is}}^{t} = 0$) during the generation.
Therefore, the effective number of steps could vary from image to image.
This modeling choice is not restrictive if one pre-specifies a reasonably large $T$.
Also, we found that such a sequence of Bernoulli random variables $\{\rvz_{\text{is}}^{t}\}$ is easier to learn compared to the geometric distribution used by AIR.
For example, AIR tends to learn very small values of $T$ (\eg, $T=2$ or $3$) in practice.
We will explain how to construct the probability distributions of these latent variables in Section \ref{sect:prior}.

\paragraph{Overall Generative Process}
The overview of our generative process is shown in Fig. \ref{fig:model}.
The sampling from the prior $p_{\theta}(\rvz^{1}, \dots, \rvz^{T})$ is like drawing one part at a step on a blank canvas for $T$ steps (with the possibility to skip some generation steps), which should produce the rough objects/scenes of the image.
The conditional sampling from the decoder $p_{\theta}(\rvx \vert \rvz^{1}, \dots, \rvz^{T})$ then generates the image by completing the fine details on the canvas.
In the following, we will explain the design choices of the individual components in detail.

\subsection{Non-parametric Structured Prior}\label{sect:prior}

As discussed above, our prior model $p_{\theta}(\rvz^{1}, \dots, \rvz^{T})$ aims at sequentially drawing the main objects/scenes of an image one part at a time for $T$ steps in total.
Apparently, parts within an image have dependencies.
For example, some background parts may have strong correlation in terms of their appearance, \eg, two parts cropped from an image of sea may highly likely be blue.
Some foreground parts may often appear together like wheels and windows of a car, sometimes even under certain relative poses.
To capture such dependencies, we employ a auto-regressive model in the latent space,
\begin{align}\label{eq:auto-regressive-prior}
    p_{\theta}(\rvz^{1}, \dots, \rvz^{T}) & = \prod_{t=1}^{T} p_{\theta}(\rvz^{t} \vert \rvz^{<t}) 
\end{align}
where $p_{\theta}(\rvz^{t} \vert \rvz^{<t})$ represents the distribution of $\rvz^{t}$ conditioned on $\{\rvz^{1}, \cdots, \rvz^{t-1}\}$.
Recall that $\rvz^{t} = [\rvz_{\text{loc}}^{t}, \rvz_{\text{id}}^{t}, \rvz_{\text{is}}^{t}]$.
We decompose the conditional distribution in Eq. (\ref{eq:auto-regressive-prior}) as below,
\begin{align}\label{eq:conditional}
    p_{\theta}(\rvz^{t} \vert \rvz^{<t}) = & \underbrace{p_{\theta}(\rvz_{\text{id}}^{t} \vert \rvc^{<t}, \rvz_{\text{loc}}^{<t})}_{\text{Appearance}} \times \underbrace{p_{\theta}(\rvz_{\text{loc}}^{t} \vert \rvc^{<t}, \rvz_{\text{loc}}^{<t})}_{\text{Location}} \nonumber \\ 
    & \times \underbrace{p_{\theta}(\rvz_{\text{is}}^{t} \vert \rvc^{<t}, \rvz_{\text{loc}}^{<t})}_{\text{Skip}},
\end{align}
where $\rvc^{<t}$ is the collection of past canvases.
We introduce the canvas in the conditioning to better explain the model.
The canvas at time step $t$ is a deterministic function of all latent variables generated so far, \ie, $\rvc^{t} = g(\rvz_{\text{loc}}^{t}, \rvz_{\text{id}}^{t}, \rvz_{\text{is}}^{t}, \rvc^{t-1})$.
We will explain the canvas update function $g$ in Section \ref{subsect:canvas_update}.
$\rvc^{0}$ is initialized to be empty.
The above three probability terms correspond to modeling the appearance of the part, the location of the part, and whether to skip the current generation step respectively.

\begin{figure}[t]
    \begin{center} 
    \includegraphics[width=\linewidth]{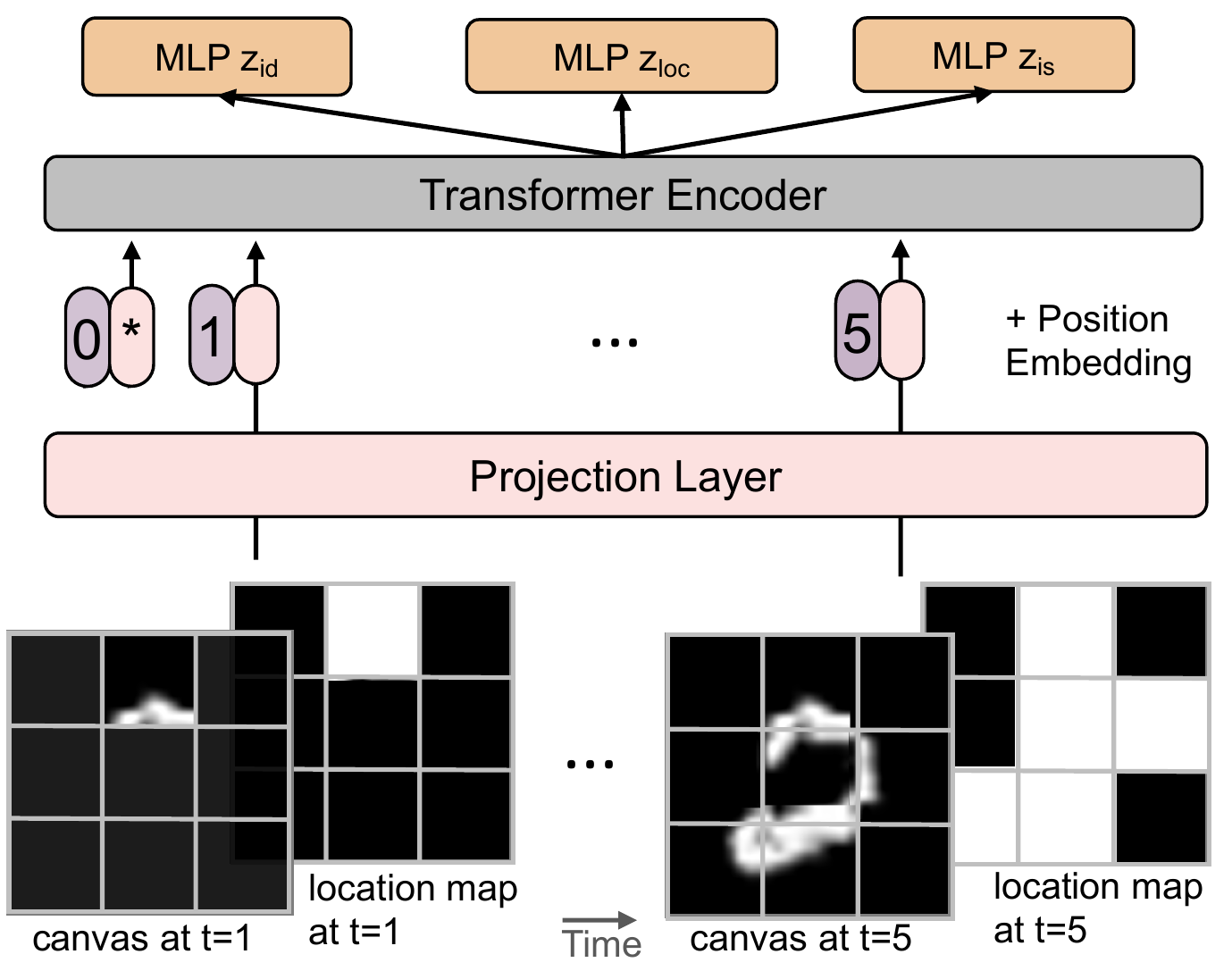} 
    \end{center}    
    \vspace{-0.5cm}
    \caption{The Transformer-based conditional probability in Eq. (\ref{eq:conditional}). The input is a sequence of 2-channel images containing the canvas and the location map.}
    \vspace{-0.4cm}
    \label{fig:transformer}
\end{figure}

\paragraph{Transformer Backbone}
To model the dependencies of all previous canvases and locations which are of variable lengths, we leverage Transformer \citep{vaswani2017attention} as the backbone.
We use the same architecture as Vision Transformer (ViT) \citep{dosovitskiy2020image}.
Specifically, at time step $t$, we feed a length-$(t-1)$ sequence of 2-channel images (each image corresponds to a time step in the past) to Transformer. 
The first channel of an image contains the latent canvas at that time step whereas the second one is a binary mask which encodes the location of the part (pixels belong to the part are set to $1$ otherwise $0$) generated at that time step.
The canvas contains the appearance information of all parts drawn so far.
The binary masks help the model aware of the locations of the recently generated parts which often provide strong clues about where to draw the next part.
We found conditioning on the previously generated locations empirically improves the performance.
Our design choice facilitates more efficient learning compared to RNNs used by DRAW and AIR, since Transformer enjoys the parallel training and is easier to optimize. 
The process is illustrated in Fig. \ref{fig:transformer}.
In the following, we explain how to construct the probabilities from the embeddings of Transformer in detail.


\subsubsection{What-to-Draw: Non-parametric Appearance Modeling for Image Parts}\label{subsect:what}

To model the appearance of individual parts, a straightforward idea is to model $p_{\theta}(\rvz_{\text{id}}^{t} \vert \rvc^{<t}, \rvz_{\text{loc}}^{<t})$ using a parametric family of distributions, \eg, Gaussians as used in DRAW and AIR.
However, such a parametric form has the limited expressiveness and hardly captures the large variation of the appearance in natural images.
We instead take a non-parametric approach by first constructing a bank of exemplar parts through clustering and then representing $\rvz_{\text{id}}^{t}$ as a categorical random variable over the choice of exemplars.

In particular, we first collect raw patches with size $K \times K$ from the training dataset through random sampling.
Then we vectorize the patches and perform K-medoids clustering to obtain the bank of exemplar parts which are the $M$ cluster centroids $B = \{\tilde{\rvx}_1, \cdots, \tilde{\rvx}_M\}$.
Therefore, $\rvz_{\text{id}}^{t}$ takes value from the possible indices of parts, \ie, $1$ to $M$.

To construct the distribution $p_{\theta}(\rvz_{\text{id}}^{t} \vert \rvc^{<t}, \rvz_{\text{loc}}^{<t})$, we first pool the embeddings of each 2-channel image returned by a projection layer and add to the positional embeddings which encode the information of time steps.
We then feed the feature to the transformer encoder which uses the self-attention to improve the embedding.
The categorical distribution $\rvz_{\text{id}}^{t}$ is output by a MLP head (a size-$64$ hidden layer and a ReLU activation function~\citep{nair2010rectified}) which takes the embedding as input. 


\subsubsection{Where-to-Draw: Location Modeling}\label{subsect:where}

Previous methods model the distribution of locations of parts via Gaussians which requires truncating the out-of-image ones.
For simplicity, we model the locations of parts by discretizing the images into a 2D grid so that a part can only center on a grid.
Therefore, $\rvz_{\text{loc}}^{t}$ is a categorical random variable taking values from the possible indices of the 2D grid.
We discretize the image so that each grid is of size $K \times K$ (\ie, same size as a part), thus disallowing parts to have overlap and greatly simplifying the pipeline.
In practice, we found learning with such a categorical distribution is easier compared to Gaussians.
We use the same embedding returned by the transformer encoder and predict the categorical distribution using a separate MLP head, which has one size-$64$ hidden layer and a ReLU activation function~\citep{nair2010rectified}.
 
\subsubsection{Whether-to-Draw and Canvas Update}\label{subsect:canvas_update}

While generating both ``what-to-draw'' $\rvz_{\text{id}}^{t}$ and ``where-to-draw'' $\rvz_{\text{loc}}^{t}$ at time step $t$, we allow the model to choose whether to draw it on the canvas by sampling a per-step Bernoulli random variable $\rvz_{\text{is}}^{t}$.
The distribution is again constructed based on a separate MLP head that takes the embedding from the transformer encoder as input.
This mechanism allows the model to have varying numbers of generation steps from image to image.
We update the canvas as below,
\begin{align}\label{eq:canvas_update}    
    \rvc^{t} & = g ( \rvz_{\text{loc}}^{t}, \rvz_{\text{id}}^{t}, \rvz_{\text{is}}^{t}, \rvc^{t-1}) \nonumber \\ 
    & = \left(\rvc^{t-1} \right)^{1 - \rvz_{\text{is}}^{t}} \left( \max \left(\rvc^{t-1}, \text{Draw}(\tilde{\rvx}_{\rvz_{\text{id}}^{t}}, \rvz_{\text{loc}}^{t}) \right) \right)^{\rvz_{\text{is}}^{t}}.
\end{align}
If $\rvz_{\text{is}}^{t} = 1$, the ``Draw'' function generates a mask (same size as the canvas) where the region specified by ``where-to-draw'' $\rvz_{\text{loc}}^{t}$ is filled by values from an exemplar part corresponding to ``what-to-draw'' $\rvz_{\text{id}}^{t}$.
The values outside this designated region are zero.
Then we obtain the new canvas via a element-wise maximum between the previous canvas and the generated mask.
If $\rvz_{\text{is}}^{t} = 0$, then we keep the canvas unchanged.
The whole update process is deterministic.

\begin{algorithm*}[t]
\caption{Heuristic Parsing of Latent Variables}
\label{alg:parsing} 
\begin{algorithmic}[1]
    \State \textbf{Input}: image $\rvx$, patch bank $B = \{\tilde{\rvx}_1, \cdots, \tilde{\rvx}_M\}$ containing $K \times K$ exemplar patches, threshold $\epsilon = 0.01$
    \State Discretize image (padding if needed) into grids so that each grid is $K \times K$ (assuming there are $T$ such grids)    
    \For {$t=1,2,\ldots,T$}   \Comment{Parsing latent variables at $t$-th step}
    \State $\rvz_\text{loc}^t$ = t
    \State $\text{loc}^t$  = Center location of the $t$-th grid 
    \State $\rvx_c$ = CropImg($\rvx$, $\text{loc}^t$) \Comment {Crop a $K \times K$ patch centered at $\text{loc}^{t}$}
    \State $\rvz_\text{id}^t$ = NearestNeighbor$(B, \rvx_c)$ \Comment{Find the index of the nearest patch to $\rvx_c$ from $B$}
    \State Cost = $\Vert \rvx_c - \tilde{\rvx}_{\rvz_\text{id}^t} \Vert - \Vert \rvx_c - \mathbf{0} \Vert$ \Comment{Cost of pasting this patch vs. an empty patch}
    \State $\rvz_{\text{is}}^{t} = 1$ \textbf{if} Cost $\le \epsilon$ \textbf{else} $0$
    \EndFor
    \State \textbf{Output}: $\rvz = \left( (\rvz_{\text{loc}}^{1}, \rvz_{\text{id}}^{1}, \rvz_{\text{is}}^{1}), \dots, (\rvz_{\text{loc}}^{T}, \rvz_{\text{id}}^{T}, \rvz_{\text{is}}^{T}) \right)$
\end{algorithmic}
\end{algorithm*}

\subsubsection{Heuristic Parsing for Pre-Training Prior}\label{sect:pretrain-prior}

Learning with a structured prior in VAEs is not easy as demonstrated by vector quantized VAEs (VQ-VAEs) \citep{oord2017neural}. 
To improve the learning, we also adopt the pre-training strategy as what VQ-VAEs do.
Before we explain details of the heuristic parsing, let us denote the general probabilistic parsing of latent variables $\rvz$ given an image $\rvx$ as $p_h(\rvz \vert \rvx)$. 
To pre-train the prior, we ideally need to,
\begin{align}\label{eq:pretrain_prior_ideal}
    \min_{\theta} \quad \KL( \int p_h(\rvz \vert \rvx) p_{\text{data}}(\rvx) \mathrm{d} \rvx \Vert p_{\theta}(\rvz)).
\end{align}
However, this is impossible since we do not know the data distribution $p_{\text{data}}(\rvx)$.
In practice, we observe samples from $p_{\text{data}}(\rvx)$ and then parse $\rvx$ to get samples of latent variables $\rvz$.
Therefore, the sampled version of the objective becomes,
\begin{align}\label{eq:pretrain_prior}
    \max_{\theta} \quad \sum_{\rvx \sim p_{\text{data}}(\rvx)} \sum_{\rvz \sim p_h(\rvz \vert \rvx)} \log p_{\theta}(\rvz).
\end{align}
Note that if the parsing is deterministic, then the objective becomes the maximum log likelihood of the observed pairs $(\rvx, \rvz)$.
In our case, we propose an effective heuristic parsing algorithm to deterministically infer the latent variables $\rvz$ given an image $\rvx$ as shown in Alg. \ref{alg:parsing}.
In other words, $p_h(\rvz \vert \rvx) = \delta(\rvz = \text{Parsing}(\rvx))$ is a delta distribution.
The idea is as follows.
We first discretize the observed image into grids.
Then for each grid, we search the nearest patch in the part bank and check whether it is closer to the observed patch than an empty patch.
If so, we use its index as the parsed value for the latent variable of the appearance.
Otherwise, we choose to skip this grid. 
Using Alg. \ref{alg:parsing}, we can collect $(\rvx, \rvz)$ pairs and pre-train the prior using Eq. (\ref{eq:pretrain_prior}).

\subsection{Encoder \& Decoder}

Similar to VAEs, we approximate the posterior by learning an encoder parameterized by $\phi$.
For simplicity, we assume the following conditionally independent encoder,
\begin{equation}
     q_\phi(\rvz \vert \rvx) = \prod_{t=1}^{T} q_\phi(\rvz_{\text{id}}^{t} \vert \rvx) q_\phi(\rvz_{\text{loc}}^{t} \vert \rvx) q_\phi(\rvz_{\text{is}}^{t} \vert \rvx).
\end{equation}
We implement the encoder as a CNN which takes the image $\rvx$ as input.
Distributions of latent variables are constructed using separate MLP heads which take the flatten feature map of the CNN as input.
Our encoder CNN consists of $5$ convolutional layers with hidden size $128$, followed by Batch Normalization~\citep{ioffe2015batch} and a ReLU activation function~\citep{nair2010rectified}.
The first $3$ convolutional layers use kernel size $3$, stride size $2$, and padding size $1$. 
The last two convolutional layers use stride $1$. 
Note that one could further improve the encoder by using a Transformer-based auto-regressive posterior which captures the dependencies across time steps.
We leave it for future exploration.

To decode an image, suppose we finish the sampling from the prior and get a canvas $\rvc^{T}$ at the last time step $T$. 
The canvas should already roughly form the objects/scenes of the image.
Then we feed the canvas to a CNN decoder to generate the final image with refined details.
Our decoder CNN consists of $2$ convolutional layers with stride $2$, and two residual blocks, followed by two transposed convolutional layers. 
All layers have hidden size $128$ followed by Batch Normalization and ReLU activations. 
In particular, we model the output image as,
\begin{align}
    p_{\theta}(\rvx \vert \rvz) = \text{OutputDistribution}(\rvx; \text{CNN}(\rvc^{T})),
\end{align}
where we choose the output distribution for pixel intensities as Bernoulli for MNIST and Omniglot datasets and as the discretized logistic mixture for CIFAR10 and CelebA following \citep{Salimans2017PixeCNN}.

\subsection{Loss Functions}

The standard objective for training a VAE model is to maximize the evidence lower bound (ELBO): 
\begin{equation}
    \mathcal{L}_{\text{ELBO}} = \E_{q(\rvz \vert \rvx)} [\log p(\rvx \vert \rvz)] - \KL( q(\rvz \vert \rvx) \Vert p(\rvz) )
\end{equation}

However, we found that if we train the model with ELBO alone, the behavior of the learned posterior deviates from our expectation, \ie, the canvas should roughly form the objects/scenes of the image. 
Therefore, we add an extra regularization term defined as 
\begin{equation}
    \mathcal{L}_{\text{reg}} = - \KL( p_h(\rvz \vert \rvx) \Vert q(\rvz \vert \rvx) ).
\end{equation}
This term could guide the posterior to behave like our heuristic parser.
Therefore, we train our model by maximizing the following full objective
\begin{equation}
    \mathcal{L} = \mathcal{L}_{\text{ELBO}} + \lambda \mathcal{L}_{\text{reg}},
\end{equation}
where $\lambda$ is the weight of the regularization term. 
We set $\lambda$ as $50$ for experiments on MNIST and Omniglot, and $500$ for experiments on CIFAR-10 and CelebA. 
The value of $\lambda$ is chosen so that the scaled regularization loss is in the same order of magnitude with ELBO. 
During training, we fix the prior for simplicity. 
Exploring learning the prior together with the encoder and the decoder is left to future work.

%% file: exp.tex
\vspace{-0.1cm}
\section{Experiments}\label{sect:exp}

We test our model on four datasets, \ie, MNIST \citep{lecun1998gradient}, Omniglot \citep{lake2015human}, CIFAR-10 \citep{Krizhevsky2009a}, and CelebA \citep{liu2015faceattributes}, from three different perspectives, \ie, image generation under full training data, image generation under low-data (\ie, less data than the original training set), and the latent space editing. 
More details on datasets can be found in Appendix.

\paragraph{Baselines} 
We compare our method with two classes of models, \ie, 1) structured image models including DRAW \citep{gregor2015draw}, AIR \citep{eslami2016attend}, PixelCNN++ \citep{Salimans2017PixeCNN}, and VQ-DRAW \citep{nichol2020vq}; 2) generic generative models including VAE, 2sVAE \citep{DBLP:conf/iclr/DaiW19}, NVAE \citep{vahdat2020nvae}, WGAN \citep{arjovsky2017wasserstein}, snGAN \citep{DBLP:journals/corr/abs-1802-05957} and WGAN-GP \citep{gulrajani2017improved}. 
For all baselines, we use the publicly released code. 
We control the number of parameters of all methods roughly the same. 
More details on hyperparameters can be found in Appendix.

\paragraph{Evaluation Metric} For all experiments, we compute the FID score \citep{Heusel2017GANsTB} and the negative log likelihood (NLL) (if applicable). 
We draw 10K samples from each model and compute the FID score \wrt 10K images in the test set. 
For evaluating NLL of VAEs, we use $50$ importance weighted samples to get a tighter bound as in \citep{DBLP:journals/corr/BurdaGS15}.
We train all models for $500$ epochs and test the checkpoint with the best validation FID score.

\begin{table}
\vspace{-0.5cm}
\centering
\resizebox{\linewidth}{!}{
    \begin{tabular}{lcccc}
        \toprule
        {\bf Method} & {\bf MNIST} & {\bf Omniglot}  & {\bf CIFAR-10} & {\bf CelebA} \\ 
        & 28$\times$28 & 28 $\times$ 28 & 32$\times$32 & 64$\times$64 \\  
        \midrule
        VAE   & 16.13           & 31.97 & 106.7 & 70 \\ 
        2sVAE* & 12.6            &  - & 72.9 & 44.4 \\ 
        NVAE  & \textbf{7.93}   & \textbf{6.84} & 55.97 & \textbf{14.74}  \\ 
        snGAN & -               & - & \textbf{14.2} & -   \\ 
        WGAN  & 10.28           & 13.99 & 54.82 & 40.29 \\ 
        WGAN GP* & -            & - & 42.18 & 30.3 \\ 

        \midrule
        PixelCNN++ & 11.38 & 9.18 & 68.00 & 72.46 \\ 
        AIR      & 482.69 & 373.93 & 673.93 & 399.41 \\
        DRAW     & 27.07 & 47.22 & 162.00 & 157.00 \\ 
        VQ-DRAW  & 19.64 & 56.03 & 80.19 & 41.87 \\ 
        Ours & \textbf{2.48} & \textbf{5.53} & \textbf{62.72} & \textbf{41.10} \\ 

        \bottomrule
    \end{tabular}
}
\caption{Comparison of sample qualities (lower FID score is better). * entries are from 2sVAE paper. - means unavailable.} 
\label{table:main_res}
\end{table}


\vspace{-0.1cm}
\subsection{Image Generation}\label{subsect:image_generation}

\begin{figure*}
    \centering
    \settowidth\rotheadsize{Sampled Canvases}
    \begin{tabular}{@{\hspace{0mm}}c@{\hspace{0.5mm}}c@{\hspace{1mm}}c@{\hspace{1mm}}c@{\hspace{1mm}}c}
        \rothead{\centering Sampled Images} &
        \includegraphics[width=0.23\textwidth,trim={0 9.5cm 0 0},clip,valign=m]{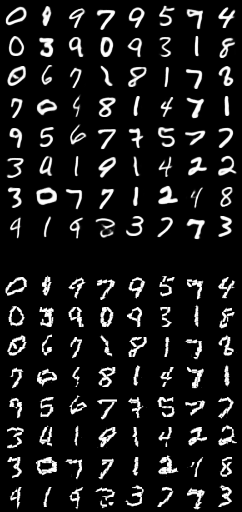} & 
        \includegraphics[width=0.23\textwidth,trim={0 9.5cm 0 0},clip,valign=m]{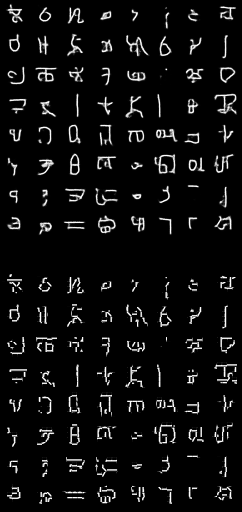} & 
        \includegraphics[width=0.23\textwidth,trim={0 10.8cm 0 0},clip,valign=m]{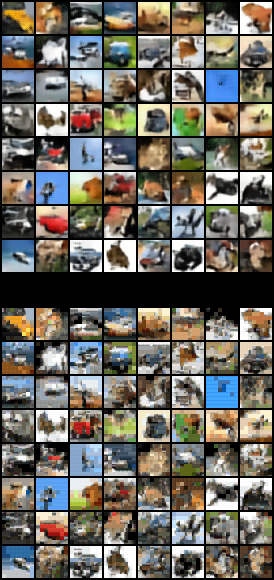} & 
        \includegraphics[width=0.23\textwidth,trim={0 21cm 0 0},clip,valign=m]{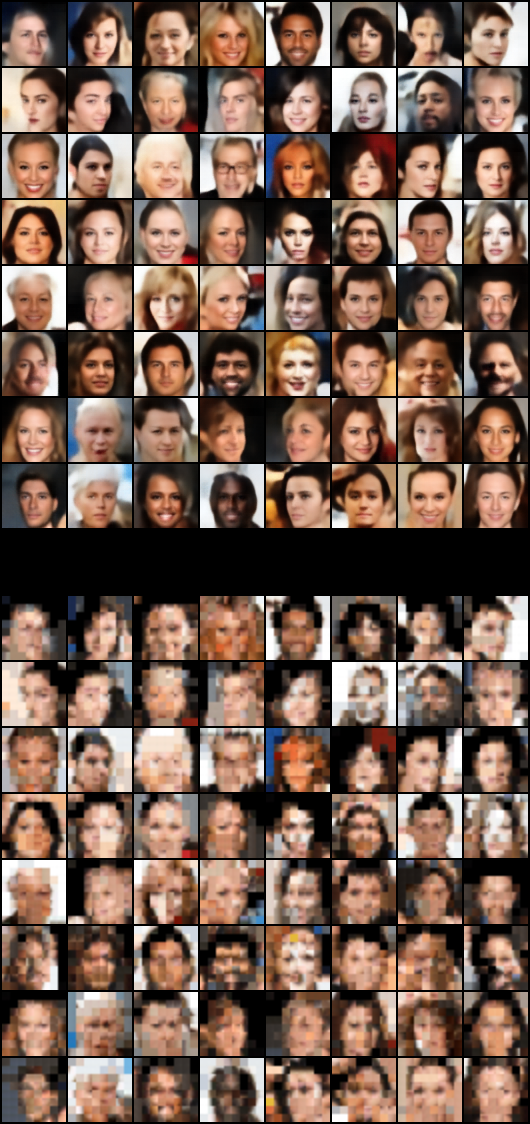} 
        \\ \addlinespace[1mm]
        \rothead{\centering Sampled Canvases} &
        \includegraphics[width=0.23\textwidth,trim={0 0 0 9.5cm},clip,valign=m]{fig/vis_sample/mnist_ours/sample_192_step_-1.png}&
        \includegraphics[width=0.23\textwidth,trim={0 0 0 9.5cm },clip,valign=m]{fig/vis_sample/omniglot_ours/sample_799_step_3.png}&
        \includegraphics[width=0.23\textwidth,trim={0  0 0 10.8cm},clip,valign=m]{fig/vis_sample/cifar_ours/sample_150_step_-1.png}&
        \includegraphics[width=0.23\textwidth,trim={0  0 0 21cm},clip,valign=m]{fig/vis_sample/celeba_ours/sample_step_129.png} \\         
        {\footnotesize  } &{\footnotesize (a) MNIST} &{\footnotesize (b) Omniglot}&{\footnotesize (c) CIFAR-10}& {\footnotesize (d) CelebA } \\
    \end{tabular}
    \vspace{-0.3cm}
    \caption{Visualization of sampled canvases ($\rvz \sim p(\rvz)$) and images ($\rvx \sim p(\rvx \vert \rvz)$) generated by our model. Same positions within the top and bottom rows correspond to one $(\rvx, \rvz)$ pair.}
    \vspace{-0.4cm}
    \label{fig:vis_sec41}
\end{figure*}

\begin{figure*}[t]
    \centering
    \settowidth\rotheadsize{Omniglot}     
    \begin{tabular}{@{\hspace{0mm}}c@{\hspace{0.5mm}}c@{\hspace{1mm}}c@{\hspace{1mm}}c@{\hspace{1mm}}c@{\hspace{1mm}}c@{\hspace{1mm}}c@{\hspace{1mm}}c@{\hspace{1mm}}c}
            \rothead{\centering Omniglot} &
        \includegraphics[width=0.23\textwidth,trim={0 9.5cm 0 0},clip,valign=m]{fig/vis_sample/omniglot_ours/sample_799_step_3.png} & 
        \includegraphics[width=0.23\textwidth,trim={0 0 0 0},clip,valign=m]{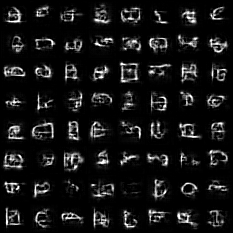} & 
        \includegraphics[width=0.23\textwidth,trim={0 0 0 0},clip,valign=m]{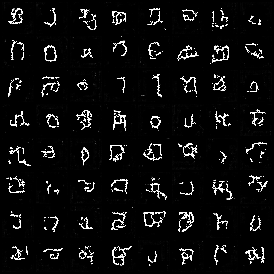} & 
        \includegraphics[width=0.23\textwidth,trim={0 0 0 0},clip,valign=m]{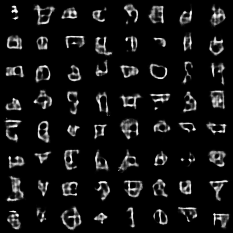} & 
            \\ \addlinespace[1mm]
            \rothead{\centering CIFAR} &
        \includegraphics[width=0.23\textwidth,trim={0 10.8cm 0 0},clip,valign=m]{fig/vis_sample/cifar_ours/sample_150_step_-1.png} & 
        \includegraphics[width=0.23\textwidth,trim={0 0 0 0},clip,valign=m]{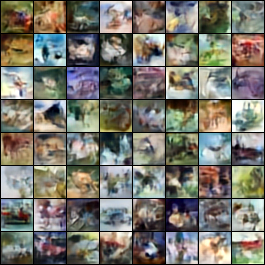} & 
        \includegraphics[width=0.23\textwidth,trim={0 0 0 0},clip,valign=m]{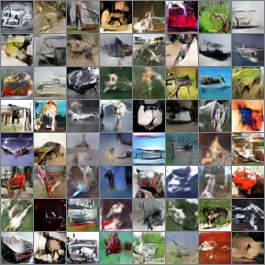} & 
        \includegraphics[width=0.23\textwidth,trim={0 0 0 0},clip,valign=m]{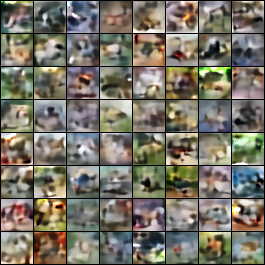} & 
            \\ \addlinespace[1mm]
            \rothead{\centering CelebA} &
        \includegraphics[width=0.23\textwidth,trim={0 21cm 0 0},clip,valign=m]{fig/vis_sample/celeba_ours/sample_step_129.png} &
        \includegraphics[width=0.23\textwidth,trim={0 0 0 0},clip,valign=m]{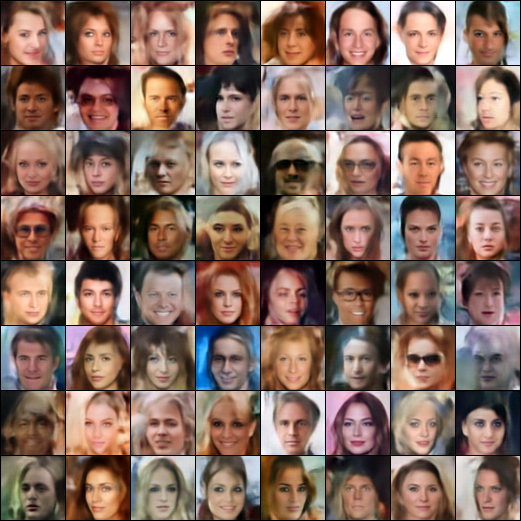} & 
        \includegraphics[width=0.23\textwidth,trim={0 0 0 0},clip,valign=m]{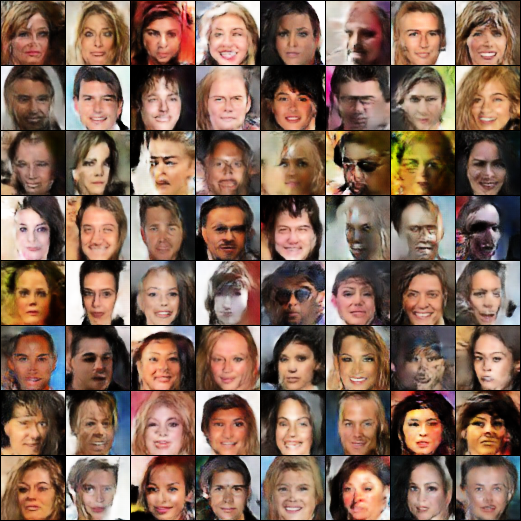} & 
        \includegraphics[width=0.23\textwidth,trim={0 0 0 0},clip,valign=m]{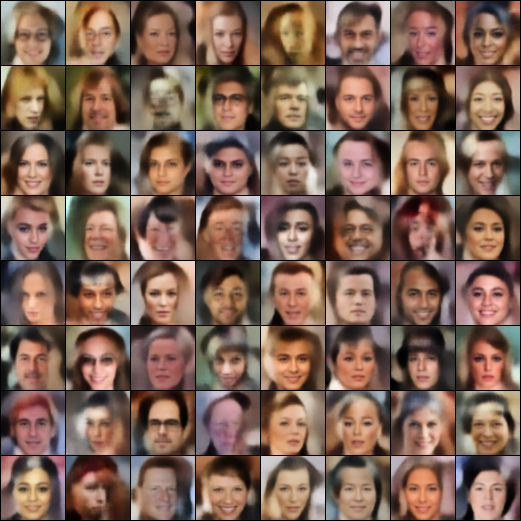} & 
        \\ 
        {\footnotesize  } 
        & {\footnotesize (a) Ours} 
        & {\footnotesize (b) VQ-DRAW} 
        & {\footnotesize (c) WGAN}
        & {\footnotesize (d) VAE}
        \\
    \end{tabular}
    \caption{Visualization of generated images from different models on different dataset.} 
    \label{fig:vis_sec61}
\end{figure*}

We first compare our model with all competitors on image generation under the full training data.
As shown in Table \ref{table:main_res}, our model significantly outperforms other structured image models in terms of the FID score. 
We also report NLLs of all models (if applicable) in Appendix. 
Our model is comparable to the state-of-the-art generic generative models like NVAE and WGAN-GP.
Moreover, our model is easier to train compared to other structured image models.
For example, while training AIR, we found it requires careful tuning to avoid a diverging loss. 
Note that AIR does not perform well on these datasets, which is likely due to their independent priors (\eg, latent locations drawn from independent Gaussians) over different time steps. 
Therefore, parts drawn at different steps tend to be randomly placed on the canvas, leading to worse FID scores.  
We provide more visualization of our generated canvases and images in Fig. \ref{fig:vis_sec41}.
It is clear that our canvases do capture the rough objects/scenes in the images.
We show more visual comparisons with other models in Fig.~\ref{fig:vis_sec61} and the Appendix.

\begin{table}[]
    \centering
    \resizebox{\linewidth}{!}{
    \begin{tabular}{lll|ccccc}
    \toprule    
    K & M & $\lambda$      & PSNR   & FID   & NLL  & BCE    & KLD   \\    
    \midrule
    \midrule
    5 & 50 & 50          & \textbf{17.71}  & \textbf{5.53}  & 129.73    & \textbf{89.95}  & 54.80 \\ 
    6 & 50 & 50          & 16.38  & 12.86 & \textbf{118.37} & 108.81 & 25.62 \\
    8 & 50 & 50          & 15.60  & 22.00 & 132.98 & 130.39 & \textbf{21.34} \\    
    \midrule
    \midrule
    5 & 50 & 50          & 17.71  & \textbf{5.53}  & 129.73    & 89.95  & 54.80 \\ 
    5 & 10 & 50          & 15.77  & 15.02 & 127.22 & 115.53 & \textbf{27.37} \\
    5 & 200 & 50         & \textbf{19.50}  & 6.08  & \textbf{115.53} & \textbf{89.94}  & 38.41 \\
    \midrule
    \midrule
    5 & 50 & 50           & 17.71  & \textbf{5.53}  & 129.73    & \textbf{89.95}  & 54.80 \\ 
    5 & 50 & 0            & 17.71  & 35.27 & 110.64 & 112.07 & \textbf{25.67} \\
    5 & 50 & 1            & 17.71  & 5.58  & \textbf{107.72} & 92.27  & 34.40 \\
    5 & 50 & 100          & 17.71  & 5.74  & 134.54 & 92.23  & 57.96 \\
    \bottomrule
    \end{tabular}
    }
\caption{Ablation study of patch size $K$ (top), part bank size $M$ (middle), and regularization $\lambda$ (bottom) on Omniglot.}
\vspace{-0.4cm}
\label{tab:ablation}
\end{table}

\begin{figure*}[t]
    \centering
     \begin{subfigure}[b]{0.33\textwidth}
         \centering
         \includegraphics[width=\textwidth]{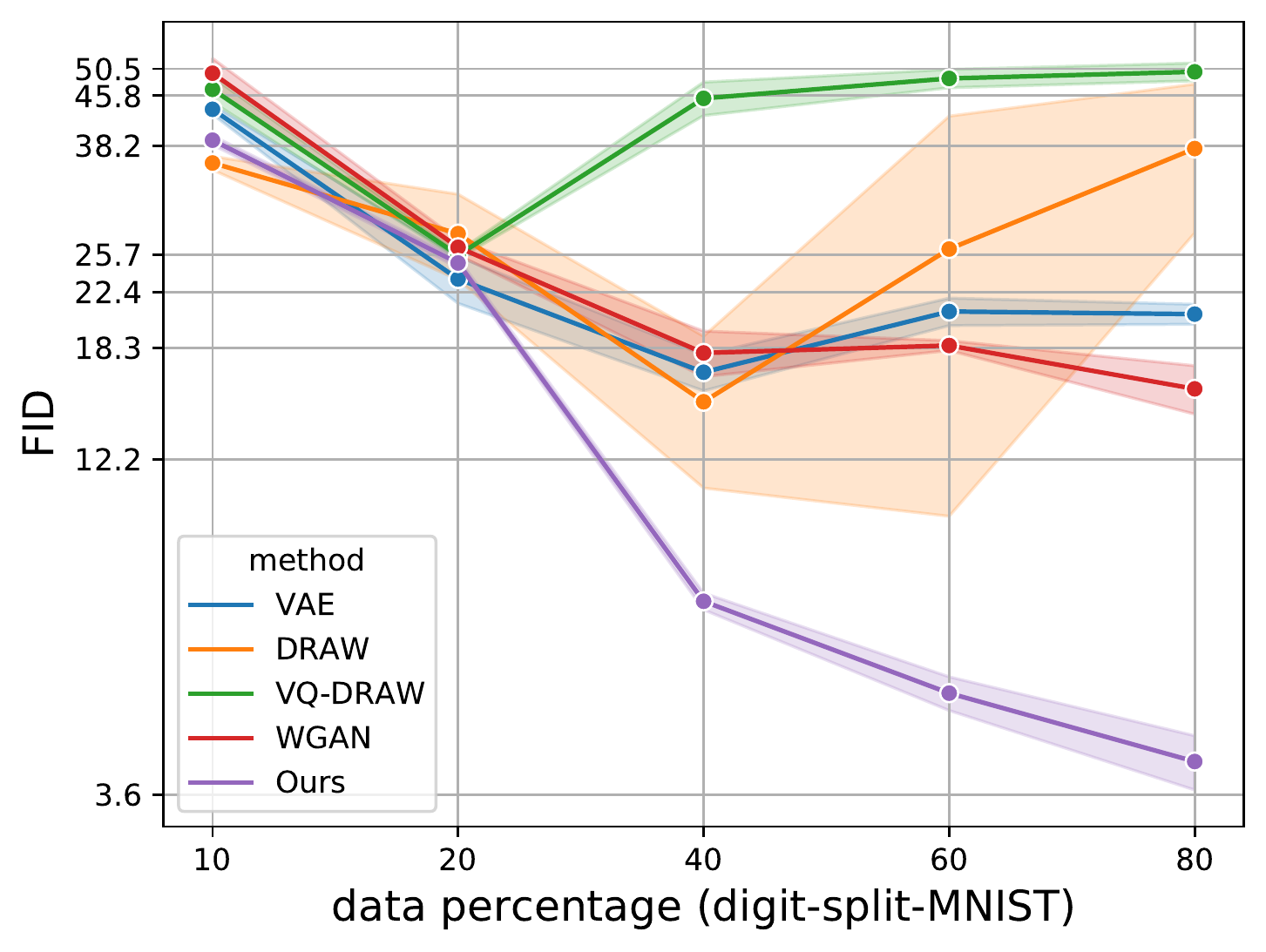}
         \vspace{-0.6cm}
         \caption{~}
         \label{fig:dmnist_fid}
     \end{subfigure}
     \hfill
     \begin{subfigure}[b]{0.33\textwidth}
         \centering
         \includegraphics[width=\textwidth]{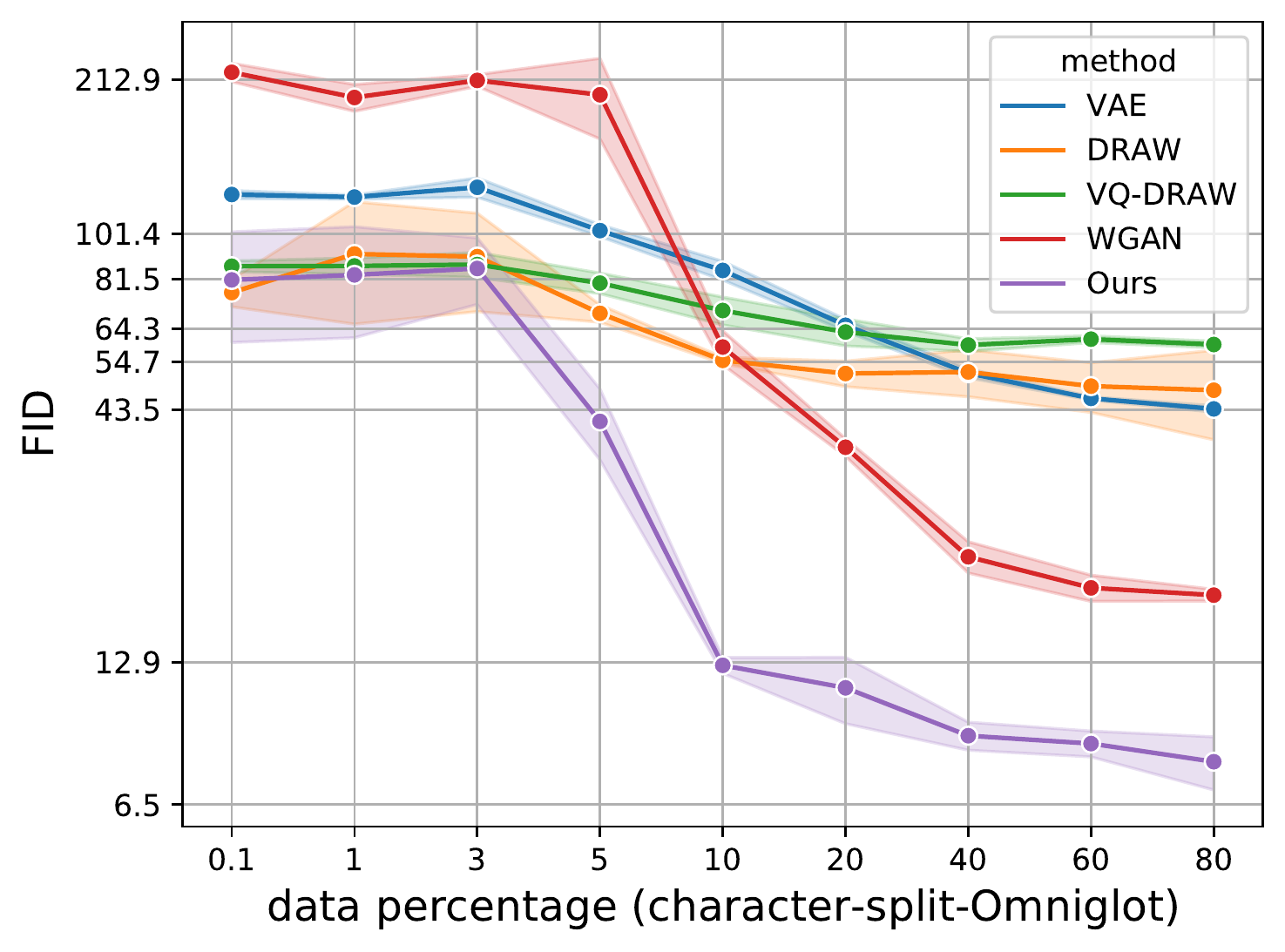}
         \vspace{-0.6cm}
         \caption{~}
         \label{fig:omni_fid}
     \end{subfigure}
     \hfill
     \begin{subfigure}[b]{0.33\textwidth}
         \centering
         \includegraphics[width=\textwidth]{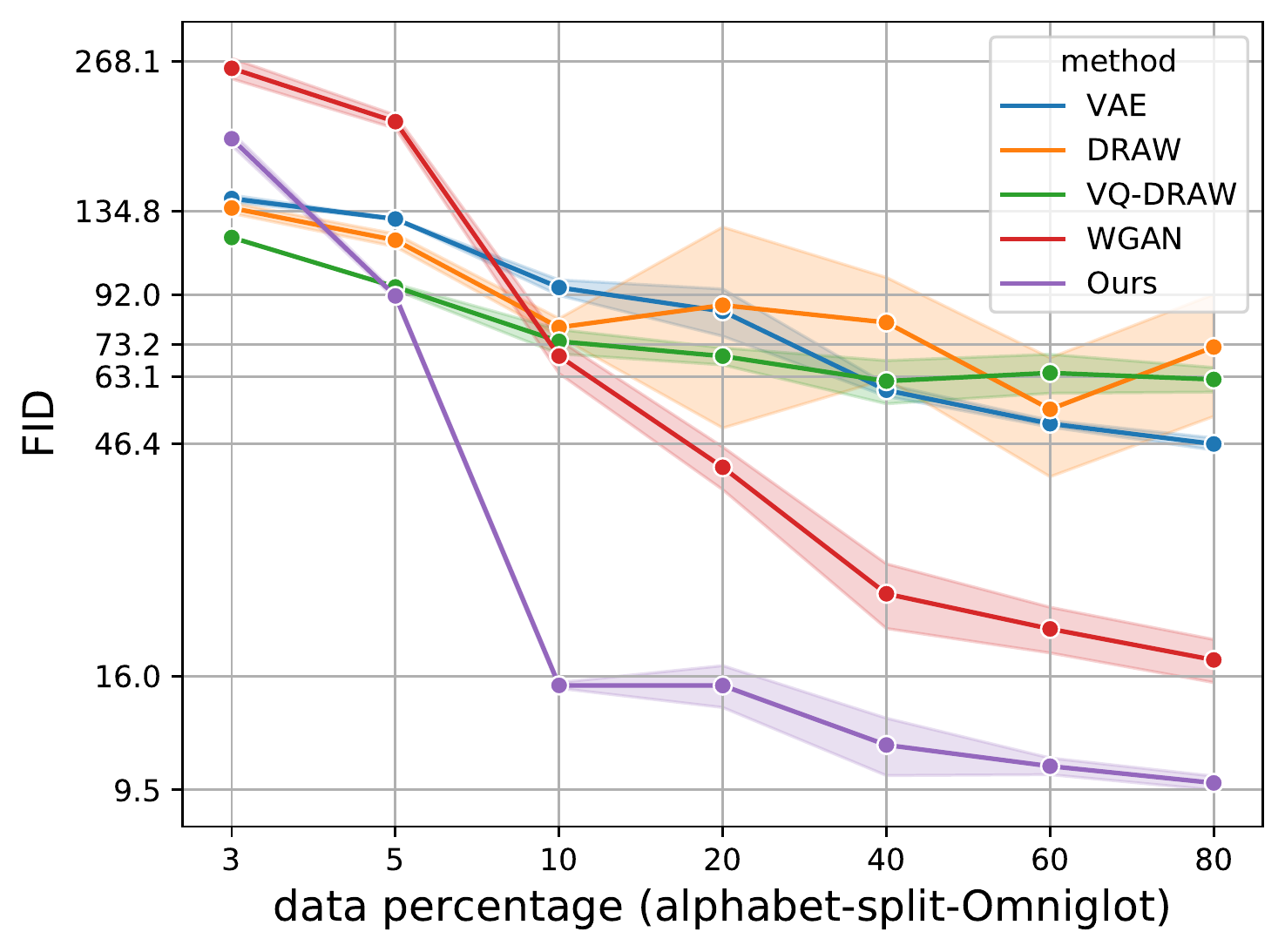}
         \vspace{-0.6cm}
         \caption{~}
         \label{fig:aomni_fid}
     \end{subfigure}\\     
     \begin{subfigure}[b]{0.33\textwidth}
         \centering
         \includegraphics[width=\textwidth]{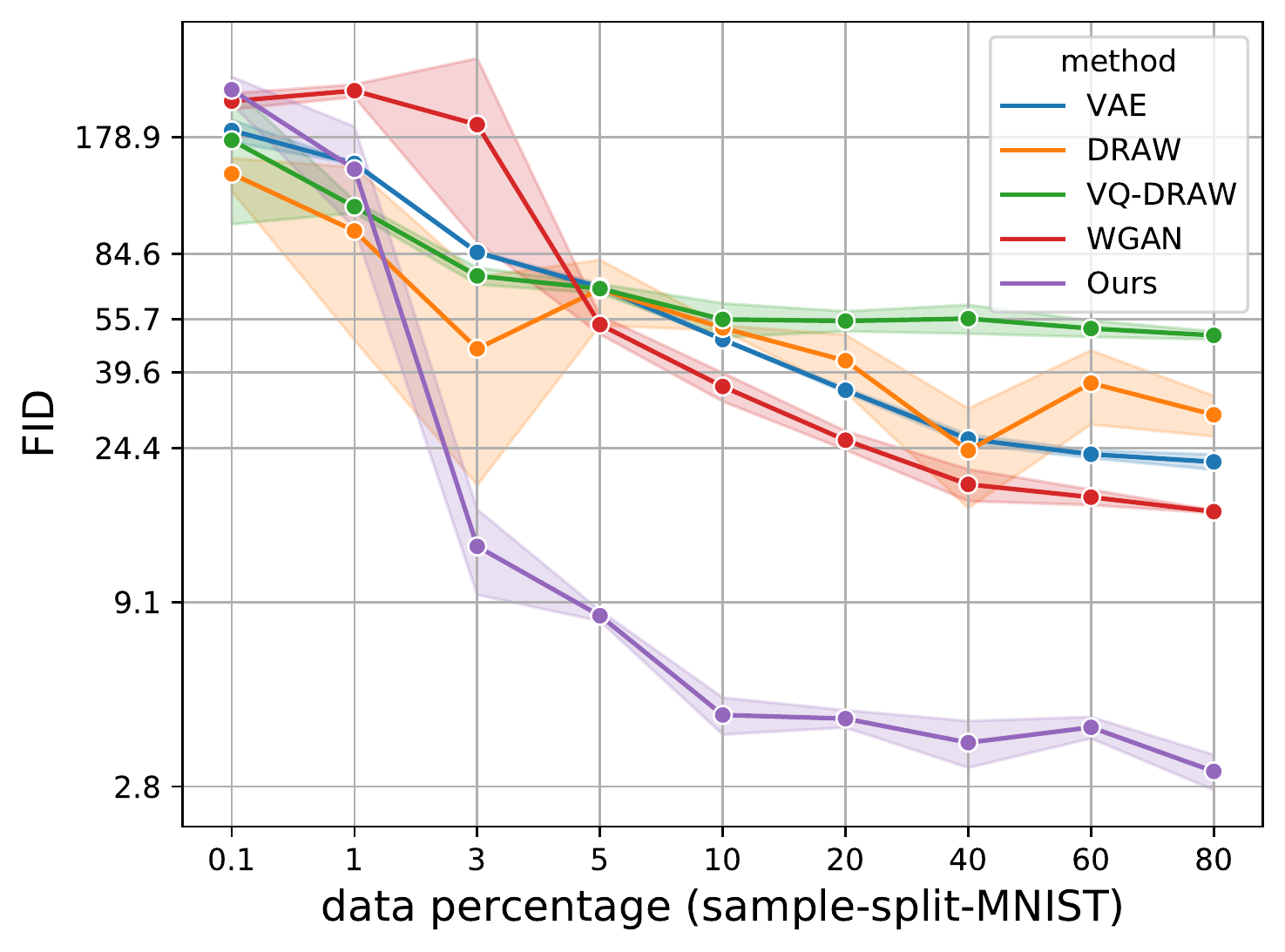}
         \vspace{-0.6cm}
         \caption{~}
         \label{fig:mnist_fid}
     \end{subfigure}
     \hfill
     \begin{subfigure}[b]{0.33\textwidth}
         \centering
         \includegraphics[width=\textwidth]{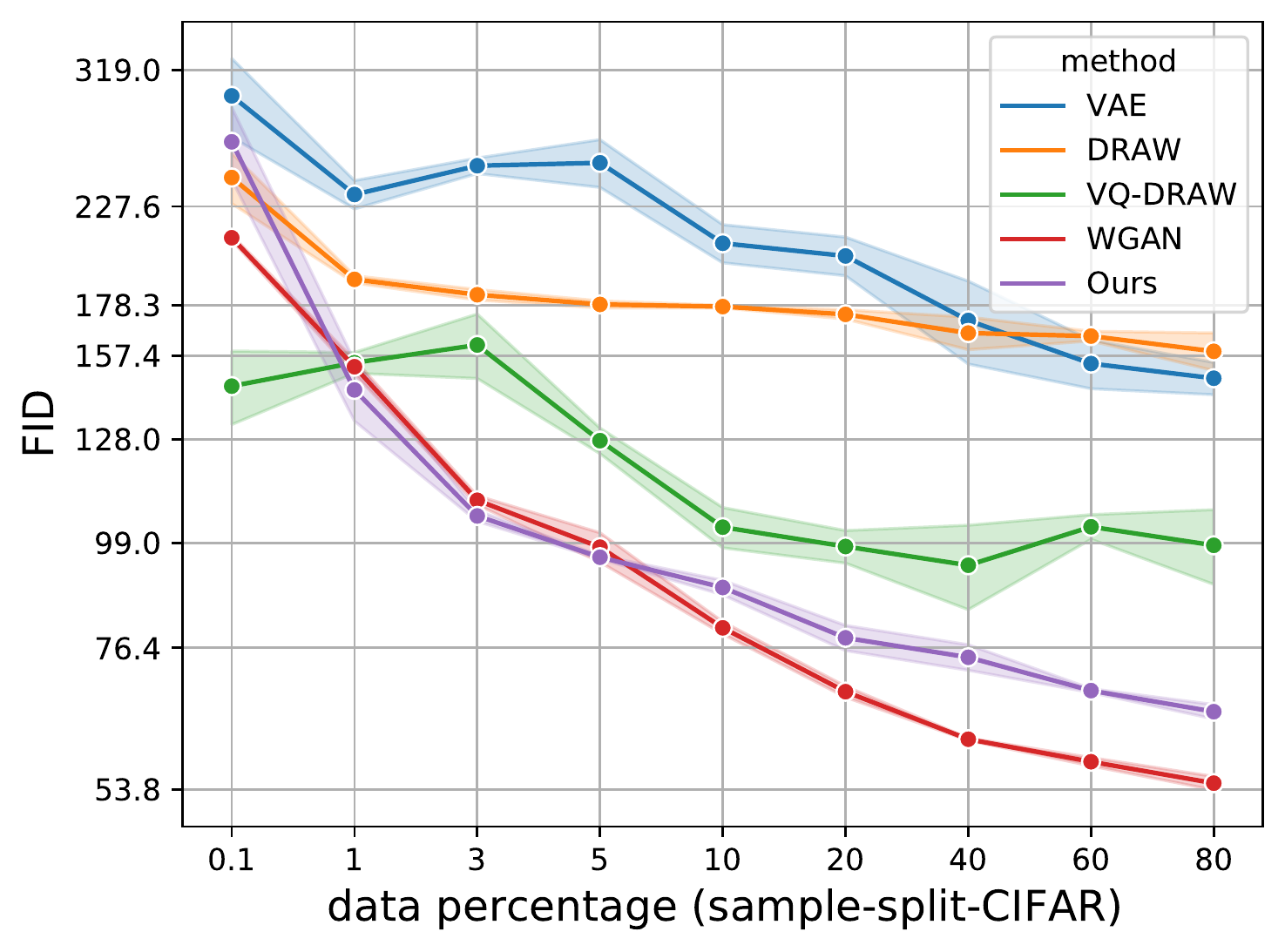}
         \vspace{-0.6cm}
         \caption{~}
         \label{fig:cifar_fid}
     \end{subfigure}
     \hfill
     \begin{subfigure}[b]{0.33\textwidth}
         \centering
         \includegraphics[width=\textwidth]{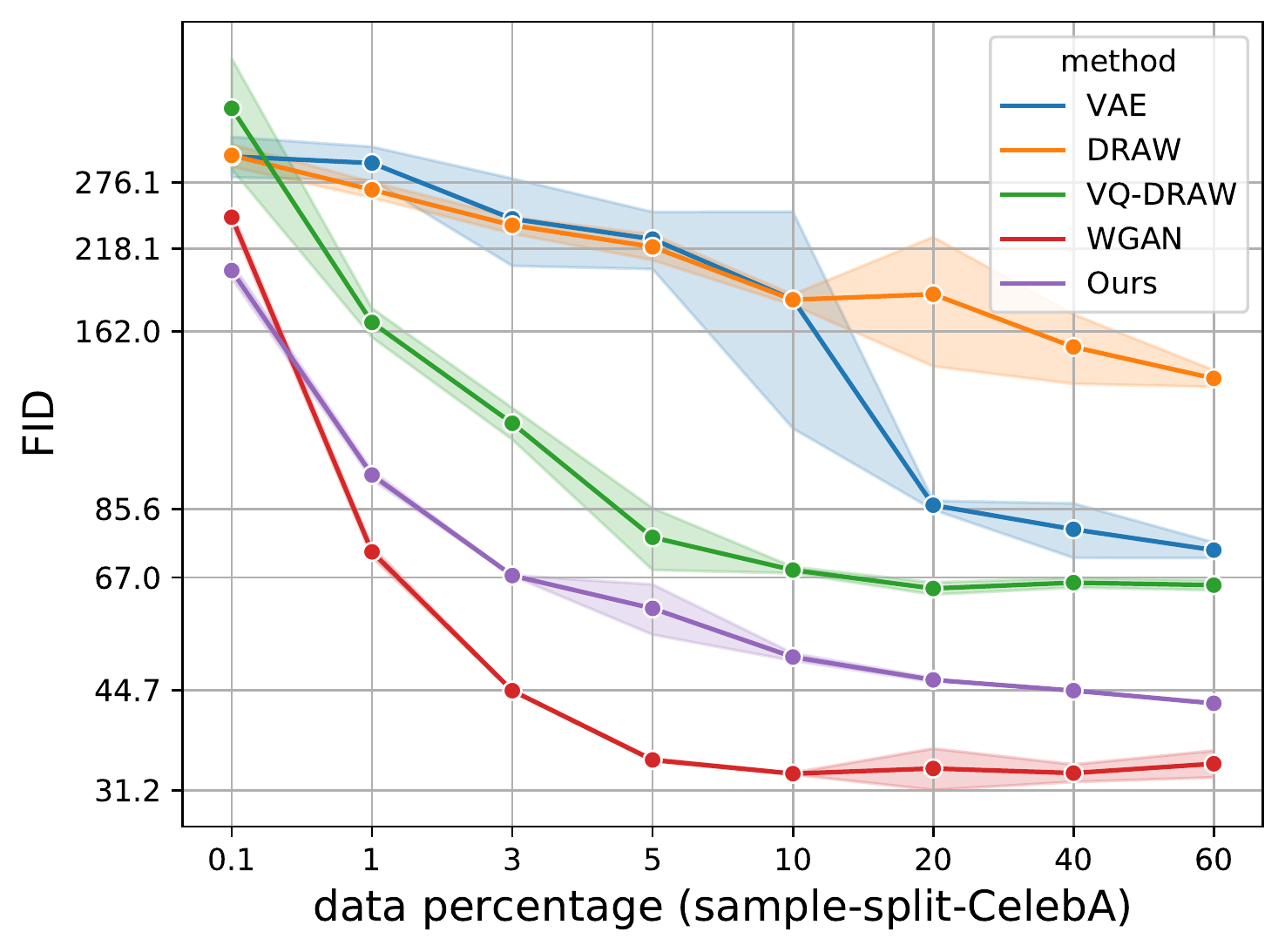}
         \vspace{-0.6cm}
         \caption{~}
         \label{fig:celeba_fid}
     \end{subfigure}     

    \vspace{-0.3cm}
    \caption{FID score vs. percentage of training data for low-data regimes. Different splits are indicated in the subcaptions. Our method consistently outperforms other structured image models. All experiments are run with three random seeds. } 
    \vspace{-0.4cm}
    \label{fig:low_data}
\end{figure*}

\vspace{-0.1cm}
\subsection{Image Generation with Low-Data}\label{subsect:exp_low_data}

In this part, we investigate our model and other baselines under low-data regimes.
The reasoning is similar to \citep{rezende2016one,lake2015human}, \ie, if our model could generate high-fidelity images with less training data, it is likely that our model better exploits the compositionality within the data.
Under each low-data setting, our model constructs the part bank only using the given limited training data, which still generalizes well to unseen images empirically. 
We group the experiments based on how we split the original dataset to create the low-data regime, \ie, split by digits in MNIST, split by characters in Omniglot, split by alphabets in Omniglot, and split by samples in Omniglot and CelebA.  
We design such splits since only MNIST has the information of digits and only Omniglot has the information of alphabets and characters.
All results are shown in Fig \ref{fig:low_data}. 

\paragraph{Split by Digits}
We split the MNIST dataset based on digits, \ie, class labels.
In particular, there are $10$ digits in total each of which occupies $10\%$ of the training data.
We train all models with different proportions of training data, \eg, $10\%, 20\%, \cdots, 80\%$ which contain digits $\{0\}, \{0,1\}, \cdots, \{0,1,2,\cdots, 7\}$ respectively.
We use the original hold-out test set to evaluate different methods which contains all $10$ digits. 
As shown in Fig. \ref{fig:dmnist_fid}, we can see that our model performs similarly with other models in the extreme cases (\ie, $10\%$ and $20\%$) and outperforms others with other percentages.
Note that the curves of DRAW and VQ-DRAW increase a bit after certain proportions which are likely due to their instable training.


\paragraph{Split by Characters}
In this setting, we split Omniglot (containing $2089$ characters in total and $11$ images per character on average) based on characters.
We randomly sample certain proportions (shown in the X-axis of Fig. \ref{fig:omni_fid}) of characters to form the new training set and validate on the rest. 
We report the final test performance on the original hold-out test set.
Since the test set contains different characters, the task is more difficult than a plain split by samples.
From Fig. \ref{fig:omni_fid}, we can see that our method consistently outperforms others under all proportions of training data.

\paragraph{Split by Alphabets}
In this setting, we split Omniglot (containing $50$ alphabets in total) based on alphabets. 
We randomly sample certain proportions (shown in the X-axis of Fig. \ref{fig:aomni_fid}) of alphabets to form the new training set and validate on the rest. 
Performances are reported on the hold-out test set.
Since the test set contains different alphabets and different alphabets in Omniglot have drastically different appearances, the task is extremely challenging than other splits.
Our method achieves the best FID scores on all proportions except $3\%$ as shown in Fig. \ref{fig:aomni_fid}.
Under the $3\%$ case, we only use $1$ alphabet for training and constructing the part bank.
The slight inferior performance of our model is likely caused by that the part bank may not be expressive enough.


\begin{figure*}[t]
    \begin{center}
        \includegraphics[width=0.98\linewidth]{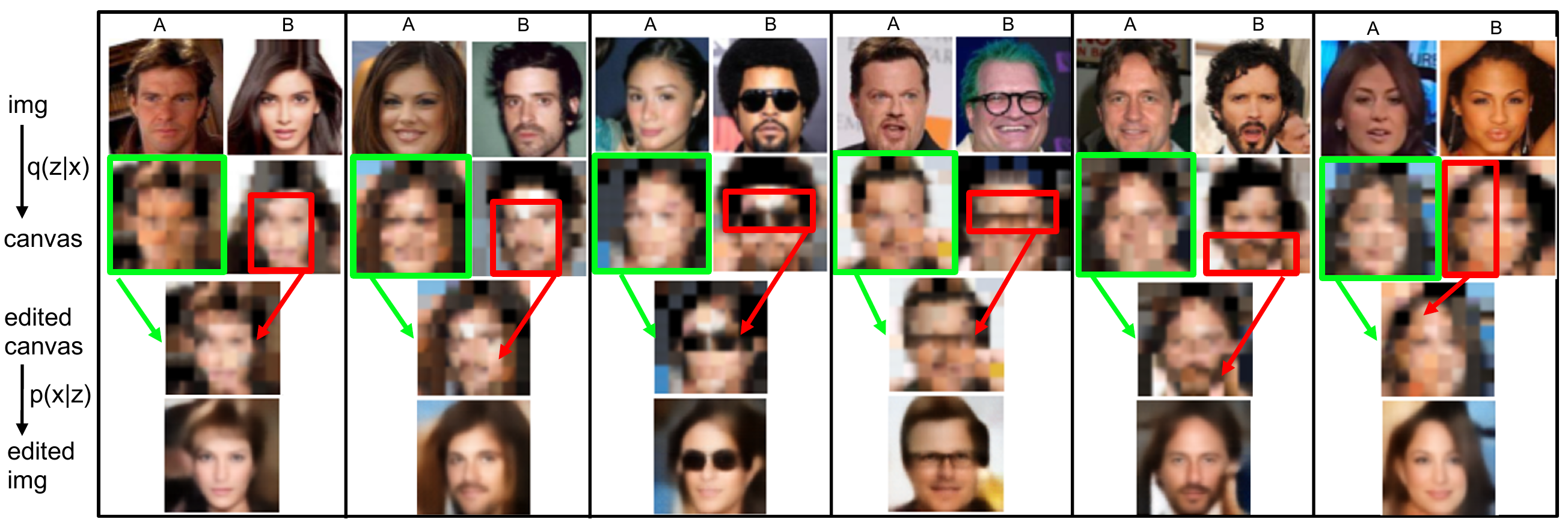} 
    \end{center} 
    \vspace{-0.5cm}
    \caption{Latent space editing. Given images A and B, we encode them to obtain the latent canvases. Then we compose a new canvas by placing a portion of canvas B on top of canvas A. Finally, we decode an image using the composed canvas.} \label{fig:vis_inter}
    \vspace{-0.2cm}
\end{figure*}

\renewcommand{\arraystretch}{1.5}

\paragraph{Split by Samples}
In this setting, we randomly sample certain proportions (shown in the X-axis of bottom subfigures in Fig. \ref{fig:low_data}) of samples to form the new training data set. 
We test this setting on MNIST, CIFAR-10, and CelebA.
It is clear that our method outperforms others under most of the proportions on MNIST.
On CIFAR-10 and CelebA, ours is superior to others except for WGAN under almost all proportions.
The gap between ours and WGAN may be attributed to the different learning principles, \ie, maximum likelihood vs. adversarial training.

\vspace{-0.2cm}
\subsection{Interpretability}\label{sect:exp:in}

In this section, we demonstrate the advantage of our interpretable latent space via interactively editing/composing the latent canvas.
In particular, given two images A and B, we first obtain their latent canvases via the encoder.
Then we manually compose a new canvas by placing a portion of the canvas B on top of the canvas A.
At last, we generate a new image by feeding this composed canvas to the decoder.
Since each part in the canvas corresponds to a local region in the sampled image, this replacement would only bring local changes, thus ensuring the spatial compositionality. 
Note that this can not be achieved by common VAEs since even a slight modification to one dimension of the latent space would typically cause global changes \wrt appearance, color, \etc.
As shown in Fig \ref{fig:vis_inter}, we can synthesize new images by placing certain semantically meaningful parts (\eg, eyeglasses, hair, beard, face) from B to A.
\renewcommand{\arraystretch}{1.5}

\vspace{-0.2cm}
\subsection{Ablation Study}
We show the ablation study of important hyperparameters on Omniglot, \ie, the patch size $K$, the size of the part bank $M$, and the regularization weight $\lambda$. 
We use the peak signal-to-noise ratio (PSNR) to measure the quality of our heuristic parsing algorithm (see more details in Appendix).
We also report the FID score, NLL, binary cross-entropy, and the KL divergence.
Results are listed in Table \ref{tab:ablation}. 
From the table, we find that the quality of our heuristic parsing improves as the part bank size $M$ increases and the patch size $K$ decreases. 
There are some inherent trade-offs on choosing the values of $M$ and $K$.
Intuitively, a large $M$ would make the learning harder since we need to deal with a higher dimensional categorical distribution, whereas a small $M$ would have inferior performance due to the lack of expressiveness.
A large $K$ is hard to learn since it requires a large $M$ to ensure the expressiveness, while a small $K$ requires more generation steps.
We hence identify a range of $K$ and $M$ based on these trade-offs and then choose the values with the best FID scores.
Moreover, adding regularization weight $\lambda > 0$ improves the performance compared to disabling it ($\lambda = 0$).
$\lambda = 50$ empirically gives the best result.

%% file: conclusion.tex
\vspace{-0.2cm}
\section{Conclusion}\label{sect:conclusion}

In this paper, we propose a structured latent variable model which generates images in a part-by-part fashion.
We construct a non-parametric distribution over the latent variables that describe the appearance of image parts.
We then propose a Transformer based auto-regressive prior to capture sequential dependencies among parts.
At last, we develop a heuristic parsing algorithm to effectively pre-train the prior.
Experiments on several challenging datasets show that our model significantly outperforms previous structured image models and is on par with state-of-the-art generic generative models.
Future work include exploring more expressive encoder, multi-scale generation, and other structured priors beyond auto-regressive models.

%% file: arxiv_appen.tex
\section{Appendix}
\subsection{Experimental Setup}\label{supp:setup}
\paragraph{Details of Datasets}
Images in MNIST and Omniglot are of size $28 \times 28$, whereas ones in CIFAR-10 are $32 \times 32$. 
We center crop images in CelebA with $148 \times 148$ bounding boxes and resize them into $64 \times 64$ following~\citep{DBLP:conf/icml/LarsenSLW16,dinh2016density}.
Omniglot has $50$ different alphabets and $2089$ characters. 
Each alphabet has $41$ characters on average and each character belongs to only one alphabet. 
There are $11$ images per character on average. 
The appearance of different alphabets vary a lot while the appearance of different characters within the same alphabet are similar. 
The original number of training images and test images are $24345$ and $8070$ respectively. 

\paragraph{Details of Baselines}

We use the open-source implementation of 
DRAW\footnote{\url{https://github.com/czm0/draw_pytorch}},
VQ-DRAW\footnote{\url{https://github.com/unixpickle/vq-draw/tree/official-release}}, 
AIR\footnote{\url{https://github.com/addtt/attend-infer-repeat-pytorch}},
WGAN\footnote{\url{https://github.com/martinarjovsky/WassersteinGAN}}, and PixelCNN++\footnote{\url{https://github.com/pclucas14/pixel-cnn-pp}}. 

For VAEs, we use an architecture with the number of parameters comparable to our model. 
We follow the hyperparameters used in the original paper if they are given. 

For DRAW, we use the same set of hyperparameters for both MNIST and Omniglot, \ie, the number of glimpses is 64, the hidden size of LSTM is 256, the latent size is 100, the read-size is $2 \times 2$, and the write-size is $5 \times 5$. 
For experiments on CIFAR-10 and CelebA, we use the number of glimpses as 64, the hidden size of LSTM as 400, the latent size as 200, the read-size as $ 5 \times 5$, and the write-size as $5 \times 5$.


Experiments on VQ-DRAW follow the default setting as the released code. 
For AIR, we set the maximum number of steps as $3$, following the setting in \citep{eslami2016attend}. 
Since AIR is originally tested on Multi-MNIST, where the number of digits, sizes of digits, and locations of the digits vary. 
To adapt AIR to datasets we use, we increase the present-probability in their prior as well as the box size and reduce the variance of the location in the prior. 
We try several groups of hyperparameters (present-probability in $\{0.01,0.6,0.8\}$, object size in $\{8,12,20\}$, std of object location in $\{1.0, 0.1, 0.001\}$, and object scale in $\{0.5,1,1.2,2\}$) for the prior.
The best hyperparameters we found are present-probability $= 0.8$, object size $= 20$, std of object location $=0.001$ for all datasets. 
The object scale is set to $0.7$ for CelebA and $1.2$ for the rest.
We also tune the weight of the KL term in their loss function.
Specifically, we try the KL weight equal to $1$ or $2$ for the ``what-to-draw" latent variable in their model.
We found $2$ is better and fix it for the rest of experiments. 

\begin{figure}
\centering 
\begin{subfigure}[t]{0.48\textwidth}
    \includegraphics[width=\textwidth,trim={0 7.45cm 0 0},clip,valign=m]{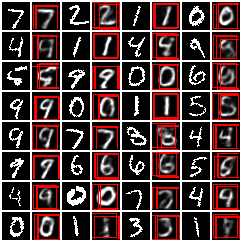}
    \caption{Reconstruction of AIR on MNIST. Each pair of images consist of the original MNIST data on the left and the reconstructed image from AIR on the right. The red boxes indicate the predicted location at each time step. We can see that AIR is able to reconstruct the original MNIST data. The model keeps refining the digits at the same location across different time steps. 
    } 
    \label{fig:vis_appendix_AIR_MNIST_a}
\end{subfigure}
\begin{subfigure}[t]{0.48\textwidth}
    \includegraphics[width=\textwidth,trim={0 6.45cm 0cm 0},clip,valign=m]{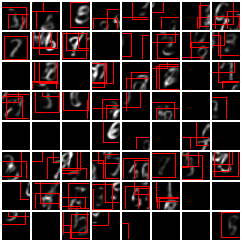} 
    \caption{Sampled images from AIR. The red boxes indicate the generated locations at each time step. The locations across time steps vary in a somewhat arbitrary manner. Moreover, the appearance of some generated digits per step are too dim to be visible. }
    \label{fig:vis_appendix_AIR_MNIST_b}
\end{subfigure}
\caption{Sampled and reconstructed images from AIR.}
\label{fig:vis_appendix_AIR_MNIST}
\end{figure}

We show generated images from AIR on MNIST in Fig. \ref{fig:vis_appendix_AIR_MNIST} to give more intuition on why AIR has a low FID score. 
From Fig. \ref{fig:vis_appendix_AIR_MNIST_a}, we can see that the reconstruction quality of AIR is reasonable.
In terms of generation, as illustrated in Fig. \ref{fig:vis_appendix_AIR_MNIST_b}, AIR fails to generate realistic images which is consistent with numbers reported in Table \ref{table:main_res:NLL}, \ie, AIR has worse NLLs on both MNIST and Omniglot datasets compared to other models. . 
One reason is that the prior of AIR is independent across time steps and is still far from being as good as the posterior (\eg, comparing Fig. \ref{fig:vis_appendix_AIR_MNIST_a} with Fig. \ref{fig:vis_appendix_AIR_MNIST_b}). 
For example, the model chooses a location to write at each time step without depending on where it wrote previously. 
Therefore, during sampling, the box location tends to jump in the canvas in an arbitrary manner. 
In our experiments, we set the variance of the object location in the prior to be small, but parts (boxes) are still placed on the canvas somewhat randomly. 
This severely degrades the sample quality since parts are not well aligned.

\paragraph{Details of Our Architecture} 

For our prior model, we use a eight layer Vision Transformer \citep{dosovitskiy2020image} with hidden size $64$ where each time step is one node and the sequence length is equal to the maximum time step. 
We use dropout with probability $0.1$ for Transformer which gives better training losses compared to the prior without dropout. 
Given a canvas at $t$, we first have a shallow CNN to extract the feature. The shallow CNN consists of two layers with hidden size 16 and a ReLU unit after the first convolutional layer. 
The canvas feature is then added to the positional embedding to serve as input feature to Transformer. 
The prior model can attend to the feature of the canvas at any time step $< t$ and then predict the ``what-to-draw" $\rvz_{\text{id}}$, ``where-to-draw" $\rvz_{\text{loc}}$, and ``whether-to-draw" $\rvz_{\text{is}}$. 
The positional embedding follows \citep{vaswani2017attention} and encodes the generation step index. 

For experiments on Omniglot and MNIST datasets, our encoder consists of four convolutional layers, while the decoder has two convolutional layers with stride $2$, two convolutional layers with stride $1$ and two transposed convolutional layers. 
All convolutional layers are followed by Batch Normalization and a ReLU activation function. 
There are three independent MLP heads to predict the approximated posteriors at $T$ time steps. 
In particular, we uniformly divide the 2D feature map into $T$ parts and feed the $t$-th feature to get the corresponding $\rvz_{\text{id}}$, $\rvz_{\text{loc}}$, and $\rvz_{\text{is}}$ at step $t$. 
We set the canvas size equal to the image size for MNIST, Omniglot, and CIFAR-10 while use a downsampled ($32 \times 32$) canvas for CelebA. 

\paragraph{Details of Training} 
We first train our prior model for $200$ epochs with batch size $64$ and learning rate $1e^{-4}$. 
After pre-training, we choose the prior model with the lowest validation loss and fix it during the training of the full model. 
The full model is trained with batch size $150$ and learning rate $1e^{-3}$. 
We use the Adam optimizer for training the model.  
The patch size and the number of parts in the part bank are $5 \times 5$ and $50$ for MNIST and Omniglot, and $4 \times 4$ and $200$ for CIFAR-10 and CelebA.
The maximum number of steps for MNIST and Omniglot is $36$ and the maximum number of steps for CIFAR-10 and CelebA is $64$.

\paragraph{Optimization}
Our latent space contains discrete distribution, which makes the loss difficult to optimize. We use gumbel-softmax~\citep{maddison2016concrete, jang2016categorical} for gradient estimation.

We can expand the KL term in our objective as follow: 
\begin{align} \small  \nonumber
    & \KL( q(\rvz \vert \rvx) \Vert p(\rvz) ) \\ \nonumber
    & = \E_{q(\rvz \vert \rvx)} \left[ \log \frac{q(\rvz \vert \rvx)}{p(\rvz)} \right] \\   \nonumber
    & = \E_{q(\rvz \vert \rvx)} \Big[ \sum_{t=1}^T 
    \log {
    q_\phi(\rvz_{\text{id}}^{t} \vert \rvx) q_\phi(\rvz_{\text{loc}}^{t} \vert \rvx) q_\phi(\rvz_{\text{is}}^{t} \vert \rvx)
    } \\ & - \log {
    p_{\theta}(\rvz_{\text{id}}^{t} \vert \rvc^{<t}, \rvz_{\text{loc}}^{<t}) 
    p_{\theta}(\rvz_{\text{loc}}^{t} \vert \rvc^{<t}, \rvz_{\text{loc}}^{<t}) 
    p_{\theta}(\rvz_{\text{is}}^{t} \vert \rvc^{<t}, \rvz_{\text{loc}}^{<t}) } \Big], 
\end{align}
where we can apply the Monte Carlo estimators. 

\subsection{More Results on Generation}
\begin{figure*}
    \centering
    \settowidth\rotheadsize{Omniglot}     
    \begin{tabular}{@{\hspace{0mm}}c@{\hspace{0.5mm}}c@{\hspace{1mm}}c@{\hspace{1mm}}c@{\hspace{1mm}}c@{\hspace{1mm}}c@{\hspace{1mm}}c@{\hspace{1mm}}c@{\hspace{1mm}}c}
        
            \rothead{\centering MNIST} &
        \includegraphics[width=0.24\textwidth,trim={0 9.5cm 0 0},clip,valign=m]{fig/vis_sample/mnist_ours/sample_192_step_-1.png} & 
        \includegraphics[width=0.24\textwidth,trim={0 0 0 0},clip,valign=m]{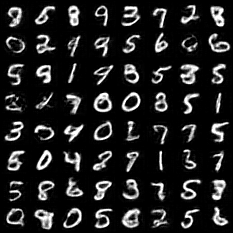} & 
        \includegraphics[width=0.24\textwidth,trim={0 0 0 0},clip,valign=m]{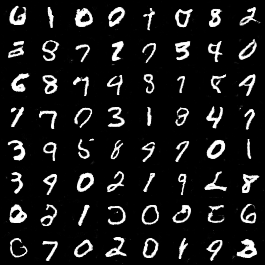} & 
        \includegraphics[width=0.24\textwidth,trim={0 0 0 0},clip,valign=m]{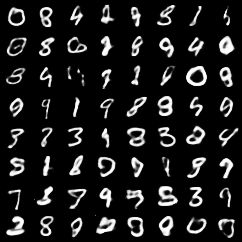} & 
            \\ \addlinespace[1mm]
            \rothead{\centering Omniglot} &
        \includegraphics[width=0.24\textwidth,trim={0 9.5cm 0 0},clip,valign=m]{fig/vis_sample/omniglot_ours/sample_799_step_3.png} & 
        \includegraphics[width=0.24\textwidth,trim={0 0 0 0},clip,valign=m]{fig/vis_sample/vqdraw/omniglot/vis.png} & 
        \includegraphics[width=0.24\textwidth,trim={0 0 0 0},clip,valign=m]{fig/vis_sample/wgan/omniglot/fake_samples_50000.png} & 
        \includegraphics[width=0.24\textwidth,trim={0 0 0 0},clip,valign=m]{fig/vis_sample/vae/omniglot/vis.png} & 
            \\ \addlinespace[1mm]
            \rothead{\centering CIFAR} &
        \includegraphics[width=0.24\textwidth,trim={0 10.8cm 0 0},clip,valign=m]{fig/vis_sample/cifar_ours/sample_150_step_-1.png} & 
        \includegraphics[width=0.24\textwidth,trim={0 0 0 0},clip,valign=m]{fig/vis_sample/vqdraw/cifar/vis.png} & 
        \includegraphics[width=0.24\textwidth,trim={0 0 0 0},clip,valign=m]{fig/vis_sample/wgan/cifar/vis.png} & 
        \includegraphics[width=0.24\textwidth,trim={0 0 0 0},clip,valign=m]{fig/vis_sample/vae/cifar/vis.png} & 
            \\ \addlinespace[1mm]
            \rothead{\centering CelebA} &
        \includegraphics[width=0.24\textwidth,trim={0 21cm 0 0},clip,valign=m]{fig/vis_sample/celeba_ours/sample_step_129.png} &
        \includegraphics[width=0.24\textwidth,trim={0 0 0 0},clip,valign=m]{fig/vis_sample/vqdraw/celeba/vis.png} & 
        \includegraphics[width=0.24\textwidth,trim={0 0 0 0},clip,valign=m]{fig/vis_sample/wgan/celeba/vis.png} & 
        \includegraphics[width=0.24\textwidth,trim={0 0 0 0},clip,valign=m]{fig/vis_sample/vae/celeba/vis.png} & 
        \\ 
        {\footnotesize  } 
        & {\footnotesize (a) Ours} 
        & {\footnotesize (b) VQ-DRAW} 
        & {\footnotesize (c) WGAN}
        & {\footnotesize (d) VAE}
        \\
    \end{tabular}
    \caption{Visualization of generated images from different models on different dataset.} 
    \label{fig:vis_sec61}
\end{figure*}

In Fig \ref{fig:vis_sec61}, we show more visualization of sampled images from different models on all datasets. 
From the figure, we can see that our method clearly outperforms VAE and VQ-DRAW and is comparable to WGAN on all datasets (even better than WGAN on Omniglot).
We also compare our model against state-of-the-art likelihood based generative models in Table \ref{table:main_res:NLL}.
We again approximate the likelihood of VAE-family of models using importance sampling with $50$ samples.
From the table, we can see that ours is comparable to other structured image models but worse than other generic models including the vanilla VAE.
This is likely caused by two facts. 
First, the construction and the dimension of the latent space between ours and other VAEs are very different which would affect the numerical values of ELBO.
Second, likelihood (or ELBO) in general is not a faithful metric that well captures the visual quality of images.

\renewcommand{\arraystretch}{1.5}

\begin{table}
\centering
\resizebox{0.49\textwidth}{!}{
    \begin{tabular}{lcccc}
        \toprule
        {\bf Method} & {\bf MNIST} & {\bf Omniglot}  & {\bf CIFAR-10} & {\bf CelebA} \\
                     & 28$\times$28 & 28 $\times$ 28 & 32$\times$32 & 64$\times$64 \\

        \midrule 
        VAE       & $\leq$ 87.00 & $\leq$ 107.90 & $\leq$ 5.70 & $\leq$ 5.76 \\
        NVAE & $\leq$ 79.60 & $\leq$ 92.80 & $\leq$ \textbf{2.91} & $\leq$  $\textbf{2.06}$ \\
        BIVA & $\leq$ \textbf{78.41} & $\leq$ \textbf{91.34} & $\leq$ 3.08 & $\leq$ 2.48   \\
        Real NVP  & 80.34 & 99.60 & 3.49 & 3.07 \\

        \midrule 
        PixelCNN++   &  79.92       & \textbf{90.82}      &  \textbf{3.10}       & \textbf{2.26} \\
        AIR          & $\leq$ 128.40  & $\leq$ 116.08 & * & * \\ 
        DRAW         & $\leq$ \textbf{66.07}  & $\leq$ 96.54 & * & *  \\ 
        Ours         & $\leq$ 98.92 & $\leq$ 129.73 & $\leq$ 5.48 & $\leq$ 5.89 \\

        \bottomrule
    \end{tabular}
}
\caption{Comparison against the state-of-the-art likelihood-based generative models. The performance is measured in bits/dimension (bpd) for all the datasets but MNIST and Omniglot in which negative log-likelihood in nats is reported (the lower the better in all cases). * entries are incomparable since they are using continues likelihood for data. } 
\label{table:main_res:NLL}
\end{table}

\subsection{More Results on Low-Data Learning}

\begin{figure*}
    \centering
    \settowidth\rotheadsize{{\scriptsize Alphabet-Split-Omniglot}}
    \begin{tabular}{@{\hspace{0mm}}c@{\hspace{0.5mm}}c@{\hspace{0.5mm}}c@{\hspace{0.5mm}}c@{\hspace{1mm}}c@{\hspace{1mm}}c@{\hspace{1mm}}c}
        
            \rothead{\centering {\scriptsize Sample-Split-MNIST}} &
        \includegraphics[width=0.32\textwidth,trim={0 4.1cm 0 0},clip,valign=m]{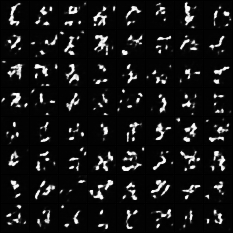} &
        \includegraphics[width=0.32\textwidth,trim={0 4.1cm 0 0},clip,valign=m]{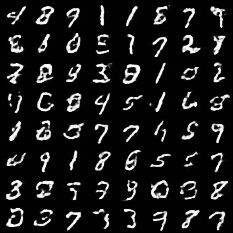} &
        \includegraphics[width=0.32\textwidth,trim={0 4.1cm 0 0},clip,valign=m]{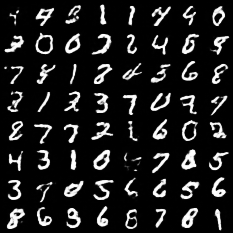} 

          \\   [-0mm]
          {\footnotesize  } 
        & {\footnotesize (a) 0.1\% } 
        & {\footnotesize (b) 1\%} 
        & {\footnotesize (c) 3\%}
            \\  [-0mm] 
            \rothead{\centering {\scriptsize Alphabet-Split-Omniglot}} &
        \includegraphics[width=0.32\textwidth,trim={0 4.1cm 0 0},clip,valign=m]{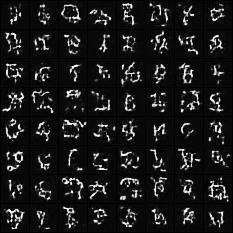} &
        \includegraphics[width=0.32\textwidth,trim={0 4.1cm 0 0},clip,valign=m]{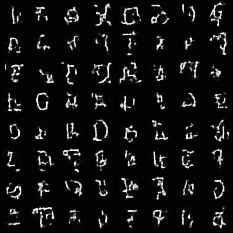} &
        \includegraphics[width=0.32\textwidth,trim={0 4.1cm 0 0},clip,valign=m]{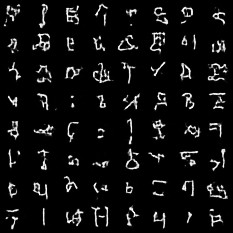} &
            \\  
          [-0mm]
          {\footnotesize  } 
        & {\footnotesize (a) 3\%}
        & {\footnotesize (b) 5\%}
        & {\footnotesize (c) 10\%}
        \\ [-0mm]
            \rothead{\centering {\scriptsize Sample-Split-CIFAR}} &
        \includegraphics[width=0.32\textwidth,trim={0 4.65cm 0 0},clip,valign=m]{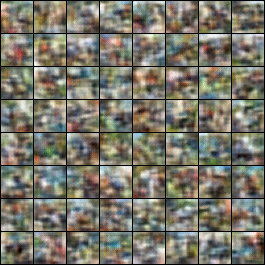} &
        \includegraphics[width=0.32\textwidth,trim={0 4.65cm 0 0},clip,valign=m]{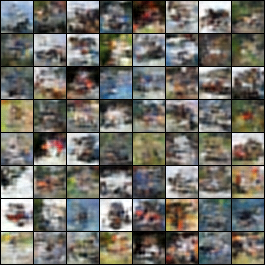} &
        \includegraphics[width=0.32\textwidth,trim={0 4.65cm 0 0},clip,valign=m]{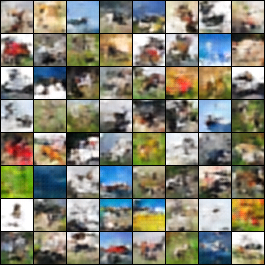} &
          \\    [-0mm]
          {\footnotesize  } 
        & {\footnotesize (a) 0.1\% } 
        & {\footnotesize (b) 1\%} 
        & {\footnotesize (c) 3\%}
            \\ [-0mm]
            \rothead{\centering {\scriptsize Sample-Split-CelebA}} &
        \includegraphics[width=0.32\textwidth,trim={0 9.3cm 0 0},clip,valign=m]{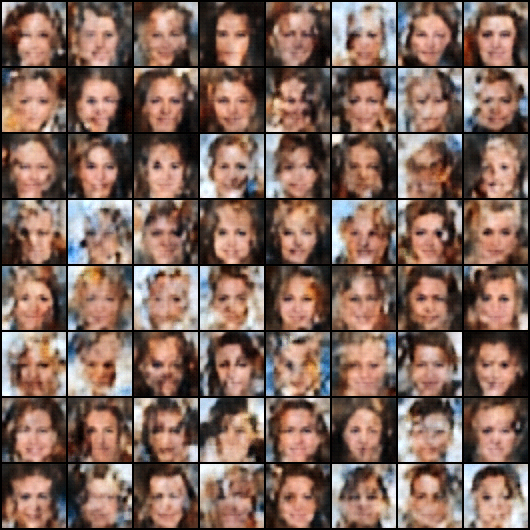} &
        \includegraphics[width=0.32\textwidth,trim={0 9.3cm 0 0},clip,valign=m]{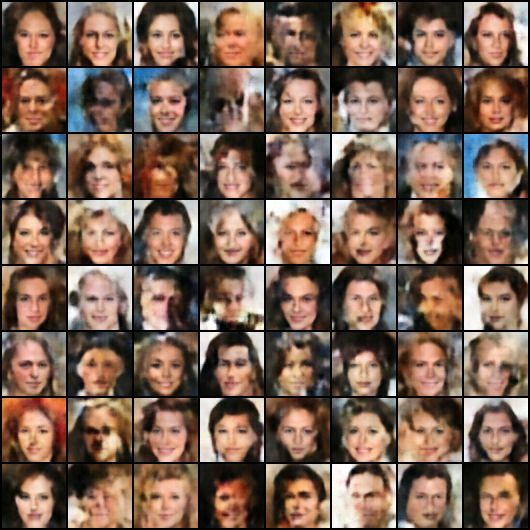} &
        \includegraphics[width=0.32\textwidth,trim={0 9.3cm 0 0},clip,valign=m]{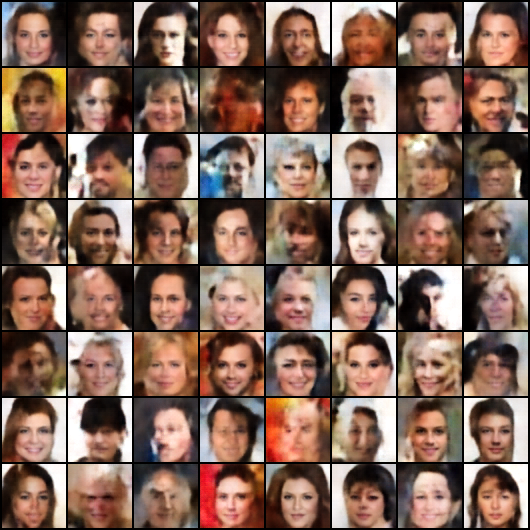} &
          \\   [-0mm]
          {\footnotesize  } 
        & {\footnotesize (a) 0.1\% } 
        & {\footnotesize (b) 1\%} 
        & {\footnotesize (c) 3\%}
        \\ [-0mm]
    \end{tabular}
    \caption{Visualization of generated images from our model on all datasets under different percentages of training data.} 
    \label{fig:vis_sec62}
\end{figure*}

In Fig \ref{fig:vis_sec62}, we show more visualization of sampled images from our model under different percentages of training data. 
As one can see from the figure, the visual quality of our samples improves as more training data is included.
Even with $0.1\%$ training data, our model could still generate plausible faces on CelebA dataset.


\subsection{Ablation Study} 
To figure out a reasonable range of values fro the patch size $K$ and the size of part bank $B$, we measure the visual quality of the output of our heuristic parsing algorithm using the peak signal-to-noise ratio (PSNR).
In particular, we paste all the selected parts returned by our heuristic parsing on a blank canvas and compute the PSNR between the original image with the created canvas. 
PSNR is computed as PSNR=$20 \log_{10}(\text{MAX}_I) - 10 \log_{10}(\text{MSE})$, where $\text{MAX}_I$ is the maximum value of the pixel intensity and $\text{MSE}$ is the mean squared error between the pasted canvas and the original image.
For the threshold $\epsilon$ used in our heuristic parsing algorithm, we set its value as $0.01$ based on the best PSNR.


\renewcommand{\arraystretch}{1.5}
